\documentclass[10pt,twocolumn,letterpaper]{article}

\usepackage{cvpr}              

\usepackage{graphicx}
\usepackage{multirow}
\usepackage{color, colortbl}
\usepackage{arydshln}
\usepackage{amsmath}

\usepackage {algpseudocode}
\usepackage{algorithmicx}
\usepackage{algorithm}
\algrenewcommand\algorithmicrequire{\textbf{Input:}}
\algrenewcommand\algorithmicensure{\textbf{Output:}}

\usepackage{subcaption}
\usepackage{wrapfig}
\usepackage{pifont}

\newcommand{\ty}[1]{\textcolor{blue}{#1}}

\def\taskname{KD}
\def\methodname{ESC}
\def\refinemethod{ESC-T}
\def\metricname{KR}
\definecolor{Gray}{gray}{0.93}	
\newcolumntype{g}{>{\columncolor{Gray}}c}

\usepackage[accsupp]{axessibility}
\usepackage{adjustbox}

\definecolor{cvprblue}{rgb}{0.21,0.49,0.74}
\usepackage[pagebackref,breaklinks,colorlinks,allcolors=cvprblue]{hyperref}

\def\paperID{8181} 
\def\confName{CVPR}
\def\confYear{2025}

\title{ESC: Erasing Space Concept for Knowledge Deletion}

\author{
{Tae-Young Lee}$^1$\thanks{Equal contribution.}
\and
{Sundong Park}$^2$\footnotemark[1]
\and
{Minwoo Jeon}$^2$\footnotemark[1]
\and
{Hyoseok Hwang}$^2$\thanks{Corresponding authors.}
\and
{Gyeong-Moon Park}$^1$\footnotemark[2] \\
\\
$^1${Korea University, Seoul, Republic of Korea} \\
$^2${Kyung Hee University, Yongin, Republic of Korea}
\and
\tt{\small{\{tylee0415, gm-park\}@korea.ac.kr}}
\\
\tt{\small{\{sundong, alsdn2530, hyoseok\}@khu.ac.kr}}
}

\begin{document}
\maketitle
\begin{abstract}
As concerns regarding privacy in deep learning continue to grow, individuals are increasingly apprehensive about the potential exploitation of their personal knowledge in trained models.
Despite several research efforts to address this, they often fail to consider the real-world demand from users for complete knowledge erasure.
Furthermore, our investigation reveals that existing methods have a risk of leaking personal knowledge through embedding features.
To address these issues, we introduce a novel concept of \textbf{K}nowledge \textbf{D}eletion (\textbf{KD}), an advanced task that considers both concerns, and provides an appropriate metric, named \textbf{K}nowledge \textbf{R}etention score (\textbf{KR}), for assessing knowledge retention in feature space.
To achieve this, we propose a novel training-free erasing approach named \textbf{E}rasing \textbf{S}pace \textbf{C}oncept (\textbf{ESC}), which restricts the important subspace for the forgetting knowledge by eliminating the relevant activations in the feature.
In addition, we suggest \textbf{ESC} with \textbf{T}raining (\textbf{ESC-T}), which uses a learnable mask to better balance the trade-off between forgetting and preserving knowledge in KD.
Our extensive experiments on various datasets and models demonstrate that our proposed methods achieve the fastest and state-of-the-art performance.
Notably, our methods are applicable to diverse forgetting scenarios, such as facial domain setting, demonstrating the generalizability of our methods.
The code is available at \href{https://github.com/KU-VGI/ESC}{https://github.com/KU-VGI/ESC}.
\end{abstract}    
\section{Introduction}
\label{sec:intro}

In response to growing concerns about data privacy, concerted efforts have been made to strengthen data protection regulations, such as General Data Protection Regulation~\cite{voigt2017gdpr}.
These regulations pose a great challenge to modern deep learning models, particularly in satisfying the requirement to erase personal information upon an individual's request.
Differential Privacy (DP)~\cite{dwork2014algorithmic} provides a prominent definition for privacy concerns, and it aims to remove individual-specific information while preserving the utility of the data for a given task.
Inspired by this concept, Machine Unlearning (MU) has recently emerged as an approach to addressing this challenge, enabling them to effectively forget selected knowledge while preserving others.

\begin{figure}[!t]
    \centering
    \includegraphics[width=0.45\textwidth]{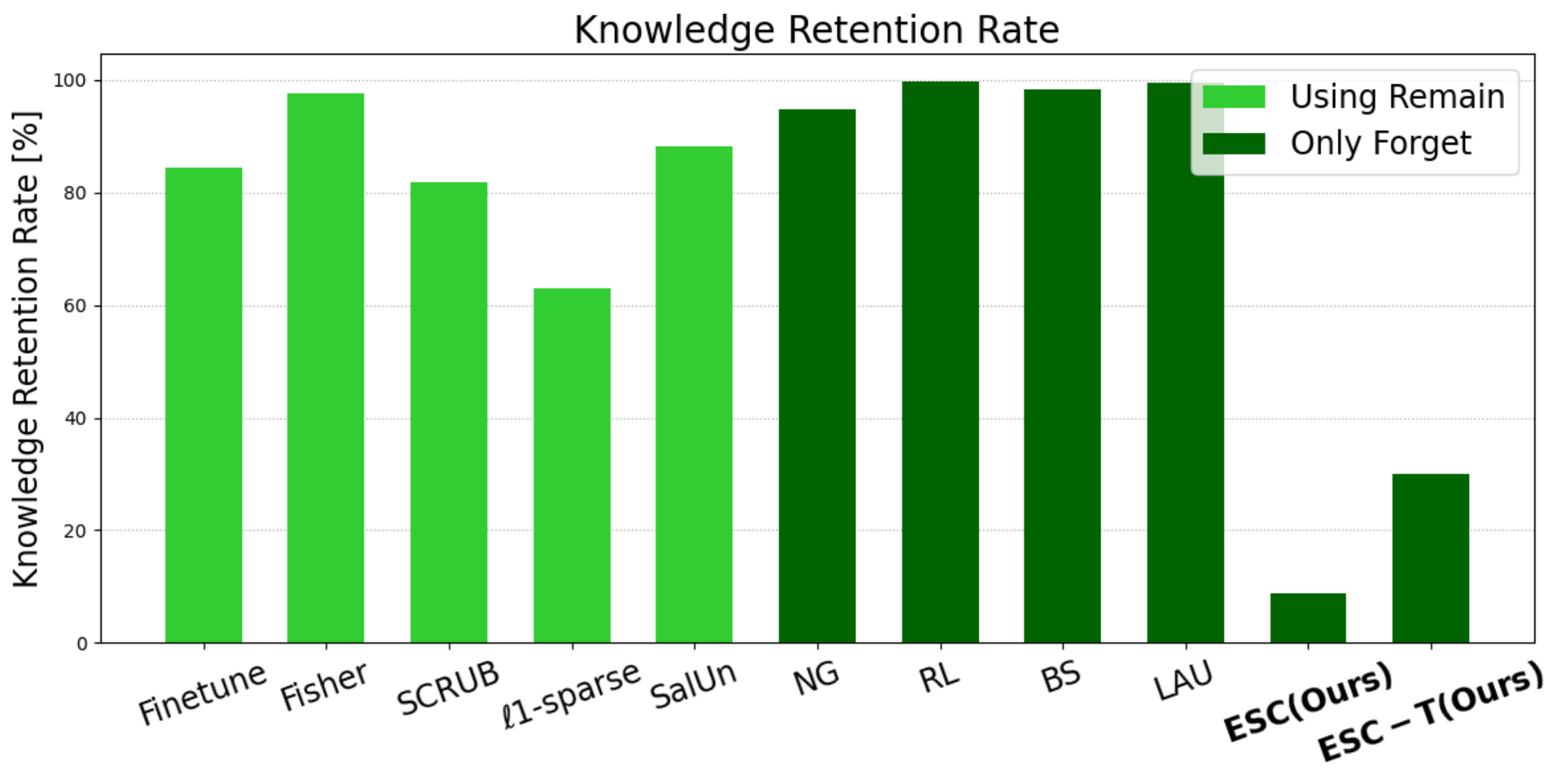}
    \vspace{-3mm}
    \caption{The recovery rate of each unlearning method using All-CNN in CIFAR-10. We conducted linear probing with each unlearned feature extractor and obtain the accuracy on the forgetting training data. Knowledge retention rate is calculated as \(acc_{f}^R/acc_{f}^O\), where \(acc_{f}^R\) is the accuracy after linear probing, and \(acc_{f}^O\) is the accuracy of the original model. While existing unlearning methods still retain the forgetting knowledge, our methods effectively eliminate it from the feature representation.}
    \label{fig:knowledge_retention}
    \vspace{-3.5mm}
\end{figure}
Previous MU approaches have largely followed privacy definitions of DP~\cite{bourtoule2021machine, golatkar2020eternal, kurmanji2023towards}, aiming to approximate a retrained model~\cite{ginart2019making} without considering the unlearned model’s state.
While this is widely adopted, it primarily focuses on data removal, overlooking practical user concerns in knowledge removal.
In real-world scenarios, users initiate removal requests with two expectations: utility-focused and privacy-focused.
Utility-focused requests seek to disable the model's ability to use the forgetting information, while privacy-focused requests aim to prevent any private information leakage, particularly against malicious attacks.
For example, if the unlearned model can still utilize the forgetting knowledge, even if it results from the remaining knowledge, users cannot fully trust the removal process.
To better meet real-world demands, we expand the scope of MU to address both utility and privacy effectively.

In addition, we point out that existing MU methods face another issue: knowledge retention in the feature. 
Our analysis of existing MU methods in Section~\ref{sec:problem_formulation} revealed that current MU approaches primarily focus on modifying the classification head while neglecting the feature extractor, which retains significant trained knowledge.
This leads to incomplete removal, as shown in Figure~\ref{fig:knowledge_retention}.
This figure indicates that retraining only a new head on top of a frozen unlearned feature extractor leads to significant performance recovery.
This suggests a significant amount of feature-level knowledge still remains.
This poses challenges for both utility and privacy.
From a utility perspective, users can still leverage forgetting knowledge via embedding features, while from a privacy perspective, malicious users can directly use these features to extract private information.

Considering the aforementioned challenges, we first define a novel \textit{\textbf{K}nowledge \textbf{D}eletion (\textbf{\taskname{}})} problem setting.
The proposed task aims to overcome the limitations of current MU methods by effectively erasing feature-level knowledge while simultaneously addressing user-specific requests. 
These requests include both utility-focused and privacy-focused needs, ultimately meeting real-world expectations for unlearning.
Moreover, we propose a new benchmark, \textit{\textbf{K}nowledge \textbf{R}etention score (\textbf{KR})}, to evaluate residual feature-level knowledge (detailed in Section~\ref{sec:knowledge_retention_score}).

To tackle \taskname{}, we propose two rapid yet powerful \taskname{} methods: \textit{\textbf{E}rasing \textbf{S}pace \textbf{C}oncept (\textbf{\methodname{}}) and \textbf{ESC} with \textbf{T}raining (\textbf{\refinemethod{}})}.
\methodname{} effectively erases the corresponding knowledge by deactivating the feature space related to forgetting data \textit{without any training process}.
\refinemethod{}, an enhanced version with lightweight optimization, uses a learnable mask to better manage the trade-off between remaining and forgetting knowledge in \taskname{}.
Through comprehensive experiments on diverse datasets and models, we demonstrate that \methodname{} and \refinemethod{} can successfully remove the forgetting knowledge from the original model, even at the feature-level.
Furthermore, our methods can be applied to various realistic \taskname{} scenarios, \eg, facial domain, and show their effectiveness.
To the best of our knowledge, our methods not only yield remarkable performances in diverse analyses but also achieve the fastest processing for \taskname{}.

Our main contributions can be summarized as follows:
\begin{itemize}
\item 
For the first time, we redefine the Machine Unlearning approach from a user-centric perspective and propose a new setting named \textit{Knowledge Deletion (\taskname{})}, which expands the scope of unlearning to feature-level knowledge.
To assess this, we introduce a novel benchmark named \textit{Knowledge Retention (KR)} score. 

\item 
To facilitate \taskname{}, we propose a novel training-free approach, \textit{Erasing Space Concept (\methodname{})}, which can eliminate the space of concept in the embedding feature space for removing the forgetting knowledge.

\item
We also propose a training-based \methodname{}, \textit{ESC with Training (\refinemethod{})}, which conducts more fine-grained removal and alleviates the trade-off between removal and preservation of the knowledge.

\item 
From the extensive experiments and analysis, we demonstrate that our proposed methods effectively erase the forgetting knowledge from the pre-trained networks in various datasets, models, and scenarios, and achieve the fastest and state-of-the-art performance in \taskname{}.

\end{itemize}
\section{Related Work}
\label{sec:related}

\paragraph{Definition of MU.}
MU is defined as a process of effectively removing the knowledge or influence of specific data from a previously trained model. 
The goal of MU is to replicate a retrain-from-scratch model~\cite{ginart2019making}, inspired by the privacy definitions in DP~\cite{dwork2014algorithmic}.
Despite its widespread use, several studies suggest that comparing unlearned models against the retrain-from-scratch model is not an appropriate evaluation method for MU.
Thudi~\etal~\cite{thudi2022necessity} theoretically demonstrate that similar model parameters can arise from different datasets.
Goel~\etal~\cite{goel2022evaluating} also suggests that retrained models can vary depending on the hyperparameters.
Therefore, existing evaluation has a risk for robust assessment of unlearned models.
Similar to Kurmanji~\cite{kurmanji2023towards}, we also consider practical requests, such as \textit{``Remove my knowledge"}, considering both utility and privacy. 
We formulate this to provide an appropriate response from the user’s perspective and extend it to the feature level. 

\vspace{-4mm}
\paragraph{MU Methods.}
MU methods can be broadly classified into two categories: exact unlearning and approximate unlearning.
Exact unlearning methods~\cite{yan2022arcane, thudi2022necessity, bourtoule2021machine, graves2021amnesiac, wu2020deltagrad, liu2021federaser, liu2022right, wang2022federated} aim to completely remove the influence of specific class or data from a model, as if it had never been used.
However, these processes still require significant computational resources.
In contrast, approximate unlearning eliminates the forgetting knowledge or data from the trained model while preserving the others.
To better preserve utility, most approaches often use the entire training dataset~\cite{golatkar2020eternal, chundawat2023can, graves2021amnesiac, tarun2023fast, fan2024salun, liu2023model, kurmanji2023towards}, which is typically suited for instance-wise unlearning.
However, in real-world scenarios, retaining sensitive data conflicts with privacy goals and legal constraints~\cite{voigt2017gdpr, rosenbaum2018data, rosenbach2018confronting, zhu2020deep}, limiting their applicability.
Recently, several methods~\cite{chen2023boundary, kim2023layer, hoang2024learn, seo2024generative} have tried to conduct unlearning only with forgetting data, and we only consider this setting in this work.
The most closely related work to ours is Kodge~\etal~\cite{kodge2024deep}, which uses Singular Value Decomposition for MU with whole datasets. 
In contrast, we consider the feature-level knowledge deletion from the user's request only with forgetting data for the first time.


\vspace{-1mm}
\section{Method}
\label{sec:method}

\vspace{-2mm}
\subsection{Preliminaries}
\label{sec:preliminaries}

\paragraph{Notations.} $\mathcal{D}=\{x_i, y_i\}^N_{i=1}$ is a training dataset consisting of the input $x_i\in X$ and its class label $y_i\in Y$, where $N$ denotes the dataset size.
We denote $\mathcal{D}_f\subseteq \mathcal{D}$ as the forgetting training data, while $\mathcal{D}_r=\mathcal{D} \backslash \mathcal{D}_f$ as the remaining training data.
Next, we denote the classification model as $f_{\theta}$ parameterized by ${\theta} $.
To illustrate our method, we further divide it into two parts: $f_{\theta}(x)=g_\phi \circ h_\psi(x) $, where $h$ is a feature extractor, $g$ is a classification head, and $\theta=\{ \phi, \psi \}$. 

\vspace{-4mm}
\paragraph{Gold Standard of MU.} 
Traditional MU aims to remove the influence of specific data or classes from a trained model based on the privacy definition of DP. For this reason, traditional MU typically considers the retrain-from-scratch model as the gold standard. 
As a result, to evaluate the efficiency of an unlearned model, most studies use privacy measures, such as accuracy or Membership Inference Attack (MIA)~\cite{carlini2022membership, choquette2021label, sablayrolles2019white, shokri2017membership, truex2019demystifying, shokri2015privacy, song2017machine}, by assessing the gap between the unlearned model and the retrained model~\cite{kurmanji2023towards, liu2023model, fan2024salun, chundawat2023can, graves2021amnesiac}. 
While MIA, which determines whether specific data was used in model training, theoretically considers the retrained model as optimal, relying solely on it as the benchmark is not always suitable for other evaluation aspects, as discussed in Section~\ref{sec:related}. 
More importantly, to meet user demands like \textit{knowledge removal}, it is essential to extend the data-centric approach to a broader context.
To address this, we redefine the MU problem to Knowledge Deletion.
In the following section, we introduce the concept of Knowledge Deletion alongside realistic user demands and propose a new optimal goal.


\subsection{Problem Formulation}
\label{sec:problem_formulation}

As already mentioned, \taskname{} reinterprets traditional MU from the user's perspective while simultaneously emphasizing the removal of knowledge at the feature level.
This section provides a detailed explanation of this setting.

\subsubsection{Extended Interpretation of MU}
We redefine the interpretation of MU to accommodate real-world knowledge removal requests, focusing on the user's perspective. 
Users interact with the deployed model and evaluate its functionality through its outputs.
For instance, if specific knowledge is removed but the model still utilizes it through the remaining knowledge, the user perceives no difference and does not consider the knowledge truly removed.
Therefore, from the utility perspective, a good knowledge-deleted model should \textit{minimize the utility of forgetting knowledge} while \textit{maximizing the utility of remaining knowledge}.

The second aspect is privacy. 
If \taskname{} focuses solely on utility, the goal could be achieved by simply modifying nodes of the final layer (\ie, head). 
However, removal requests are not restricted to the output alone; they must also ensure the protection of privacy against arbitrary attacks such as MIA. 
In \taskname{}, MIA is evaluated based on the retrain-from-scratch model, similar to previous works, as the retrained model is theoretically optimal from the perspective of membership.

\subsubsection{Knowledge Deletion.}
In this section, we highlight a key limitation of existing Machine Unlearning (MU) methods: the lack of unlearning at the feature level.
As illustrated in Figure~\ref{fig:knowledge_retention}, the knowledge retained in features indicates that the forgetting knowledge has not been fully removed, making it highly vulnerable to knowledge leakage through its outputs.
Motivated by these findings, we expand the scope of MU to the removal of feature-level knowledge.
Features capture the most representative form of learned knowledge in a trained model.
For instance, in Self-Supervised Learning (SSL)~\cite{chen2020simple, he2022masked, meila2021clip}, the quality of the extracted features is used to evaluate how effectively the model has learned from the given data, emphasizing their role as a core representation of learned knowledge.
Consequently, KD aims to effectively remove information at the feature level to ensure comprehensive knowledge removal requests in the real world.

\subsubsection{Knowledge Retention Score}
\label{sec:knowledge_retention_score}

In KD, our goal is to remove feature-level knowledge from the unlearned model, necessitating an appropriate knowledge measure in feature for evaluating the model after KD. 
However, there are no metrics for this. 
Most existing MU studies rely on metrics such as accuracy or MIA.
In addition, some studies, like Chen~\etal~\cite{chen2023boundary}, analyzed model activations using Grad-CAM~\cite{selvaraju2017grad}, or employed distance-based analyses, including activation distance~\cite{golatkar2020eternal, golatkar2021mixed}, Zero Retrain Forgetting~(ZRF)~\cite{chundawat2023can}, and JS-distance~\cite{chundawat2023can}, as well as probability-based methods like distribution of the entropy~\cite{chen2023boundary, hoang2024learn}.
Despite these efforts, all of these approaches are based on the model's output, like logit, and do not capture the feature-level information.
Although there has been an attempt to directly evaluate the model by measuring weight differences, such as layer-wise distance~\cite{tarun2023fast}, this method does not effectively determine whether the model has truly forgotten, as mentioned in Section~\ref{sec:related}.

Therefore, inspired by the Self-Supervised Learning (SSL) approach, we propose a new benchmark, named \textit{\textbf{K}nowledge \textbf{R}etention Score (\textbf{KR})}, to adequately evaluate knowledge retention at the feature level.
\metricname{} is specifically designed to measure the extent of knowledge retention in the feature extractor.
\metricname{} performs linear probing~\cite{kornblith2019better} using feature extractor after \taskname{}, and then measures the accuracy for both forgetting and remaining data.
We interpret this similarly to utility, as mentioned above.

\subsection{Erasing Space Concept}
\label{sec:esc}

\begin{figure}[!t]
    \centering
    \includegraphics[width=\columnwidth]{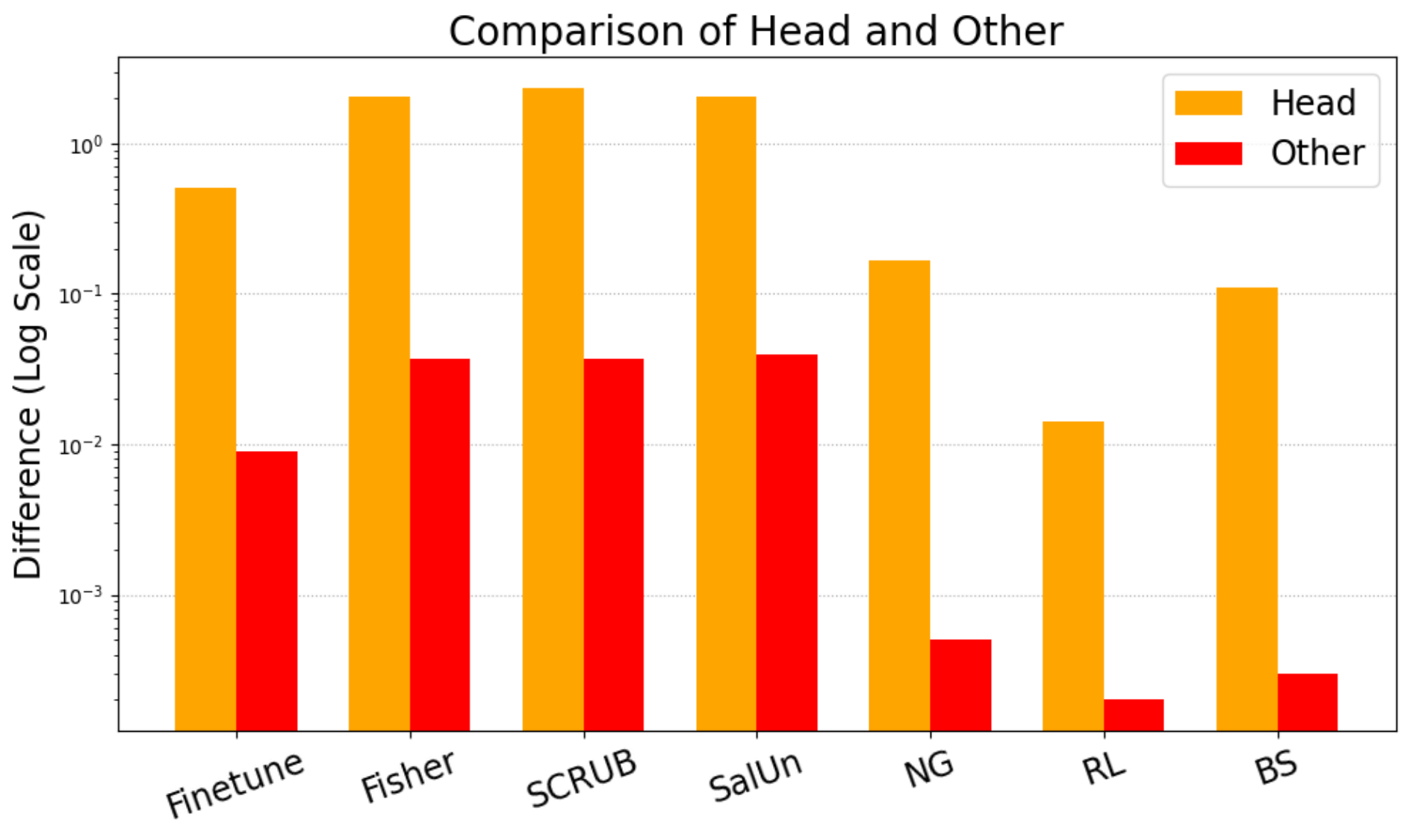}
    \vspace{-7mm}
    \caption{Using CIFAR-10~\cite{krizhevsky2009learning} with All-CNN~\cite{springenberg2014striving}, we measure weight difference between the original model and unlearned model. We calculate the difference as $\gamma\cdot\Vert\theta^l_{ori}-\theta^l_{ul}\Vert/\Vert\theta^l_{ori}$, where $\theta$ means weight of model and $\gamma$ is a scaling factor that reflects how the change in a specific layer contributes to the overall change in the model. The change in the head is dominant over the change in the entire model for all MU methods. This implies that most models focus on the head when performing MU.}
    \label{fig:weight_diff}
    \vspace{-3mm}
\end{figure}
\paragraph{Why existing works cannot unlearn at the feature level?}
Most existing MU methods develop unlearning methods based on Negative Gradient or Random Label approaches, and perform end-to-end training using these forgetting losses. 
These approaches rely on a logit-based loss function, which makes it easy for the model to find a shortcut solution by focusing on the classification head. 
In Figure~\ref{fig:weight_diff}, we experimentally analyze this by measuring the weight difference between the unlearned model and the original model. 
The results indicate that existing methods are biased toward perturbing at the head.
Consequently, most MU methods fail to effectively unlearn the encoder, where most of the learned knowledge resides.
Therefore, to achieve a more fundamental removal of knowledge, it is essential to directly erase feature-level knowledge.

\begin{figure}[!t]
    \centering
    \includegraphics[width=\columnwidth]{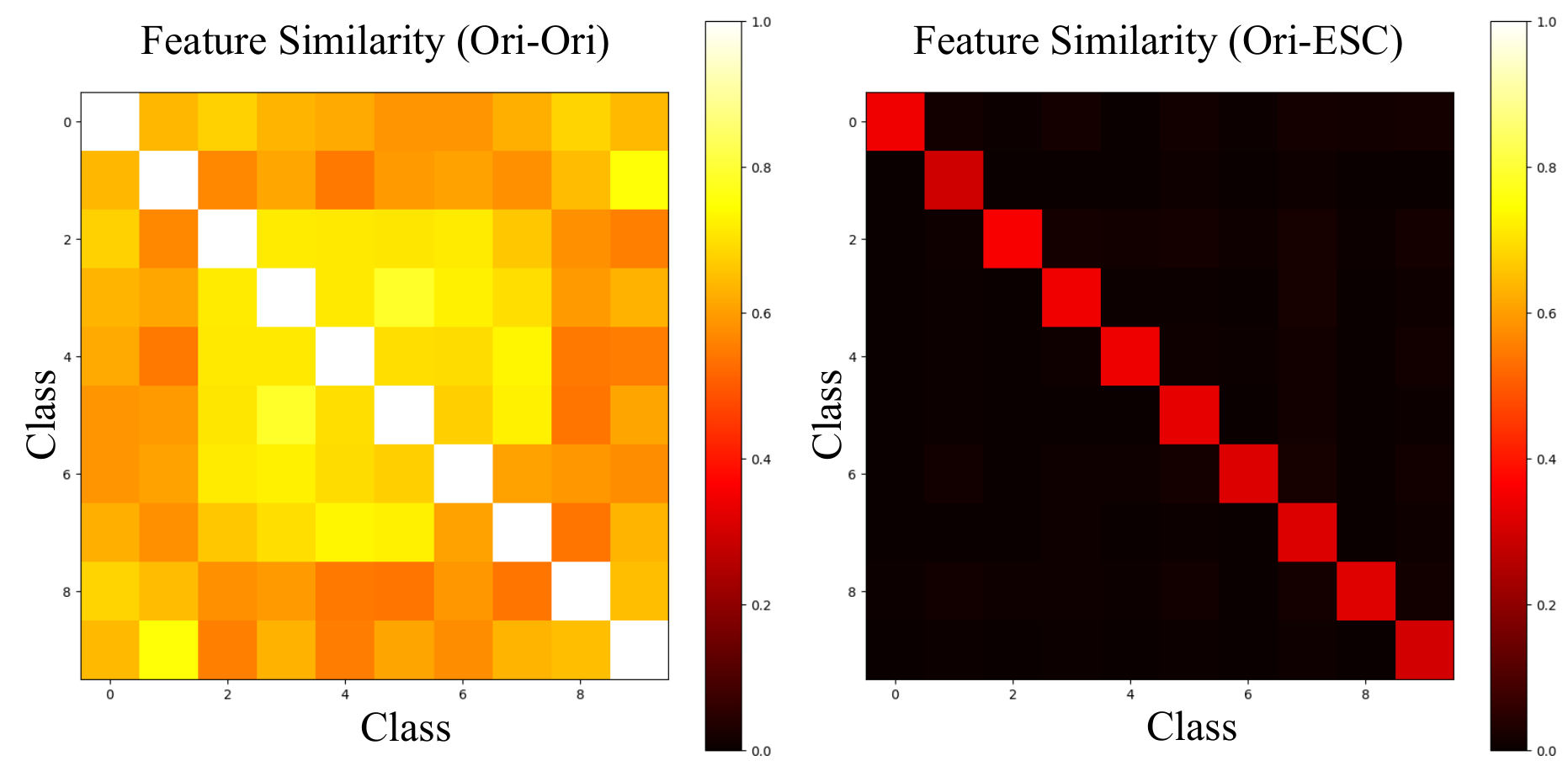}
    \vspace{-3mm}
    \caption{We calculate the cosine similarity between features for each class. (Left) Cosine similarity between the original features. (Right) Cosine similarity between the original features and the ESC features, where the principal component has been removed from the original features.}
    \label{fig:feasibility}
    \vspace{-3mm}
\end{figure}
\paragraph{Motivation.}
Our ultimate goal is to make it impossible for both users and malicious actors to extract any forgetting knowledge from the model by directly removing the knowledge in the feature space while preserving the remaining knowledge. 
To achieve this, we need to decompose the forgetting knowledge from others and remove the model's capability, \ie, activation. 
Inspired by Gu~\etal~\cite{gu2023preserving}, we can expect that we can decompose and deactivate the feature by using only partial principal directions given by Singular Value Decomposition (SVD). 
In Figure~\ref{fig:feasibility}, we conducted a toy experiment. 
We calculate the cosine similarity of features between each class. 
On the left side, the results show that the original features are highly entangled between classes, e.g., the minimum value of similarity is greater than 0.5.
On the right side, the ESC features by our method are clearly disentangled from the features of other classes (details about the ESC features are provided in the ESC section).
Furthermore, they are no longer aligned with the original features of the same class, which contain the learned knowledge, with all similarity values falling below 0.35.
Building on this insight, we introduce \textbf{E}rasing \textbf{S}pace \textbf{C}oncept named \textbf{\methodname{}}, which can remove the forgetting knowledge by deactivating in the feature space.

\begin{figure*}[!t]
    \centering
    \includegraphics[width=\textwidth]{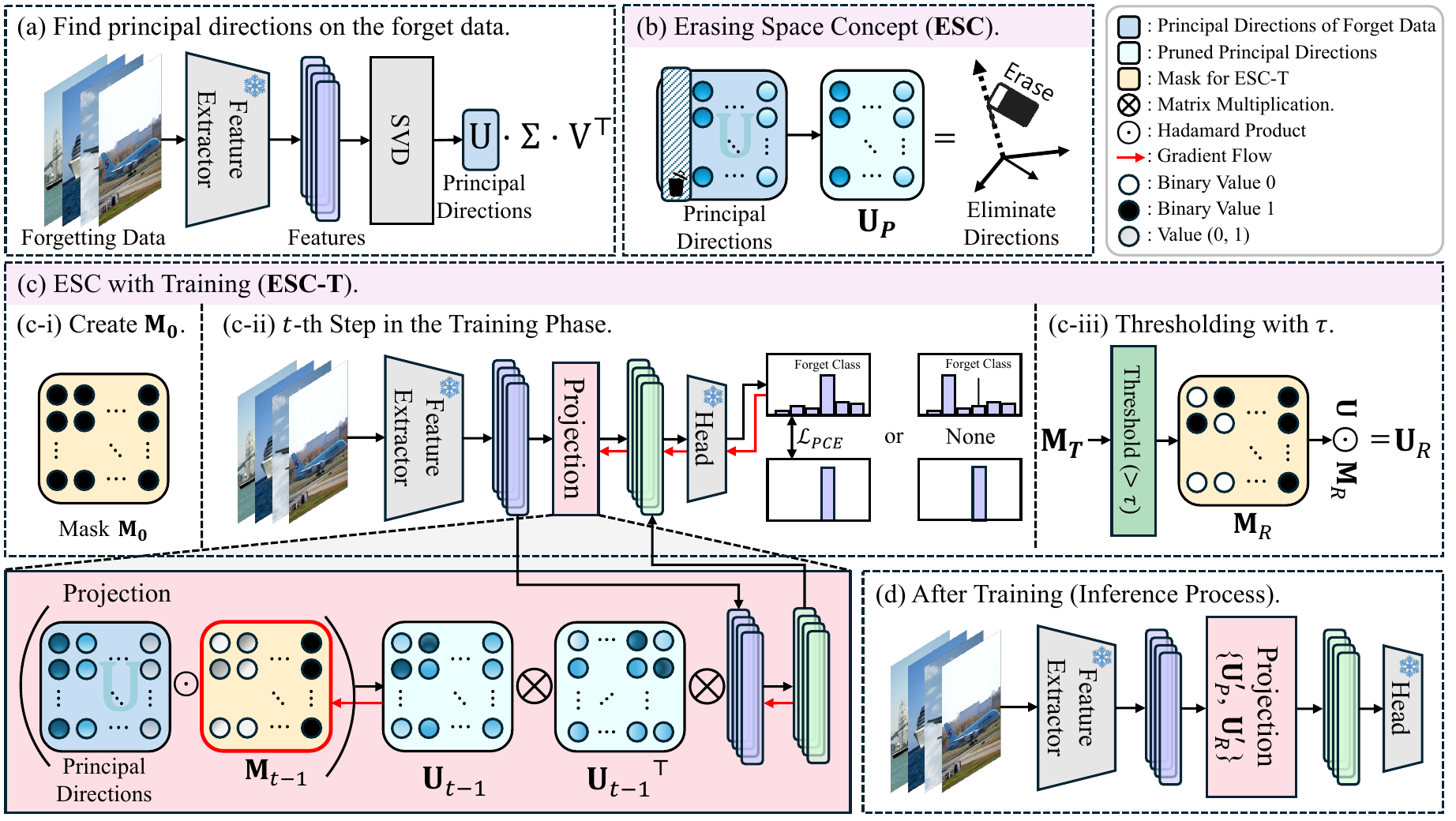}
    \caption{An overview of our methods.
    (a) We start with extracting the principal directions~($\mathbf{U}$) of embedding features from the forgetting data using SVD.
    (b) \methodname{} erases crucial directions from $\mathbf{U}$, and we can get the pruned principal directions $\mathbf{U}_P$.
    (c) \refinemethod{} enhances \methodname{} by incorporating our forgetting loss, $\mathcal{L}_{PCE}$, refining the erasure process to eliminate only important elements rather than entire directions.
    \refinemethod{} yields the refined principal directions $\mathbf{U}_R$, which improve the trade-off forgetting and preservation.
    (d) During the inference phase, we project the extracted features onto the subspace formed by $\mathbf{U}_P$ or $\mathbf{U}_R$ from each method.}
    \label{fig:main}
    \vspace{-2mm}
\end{figure*}

\vspace{-2mm}
\paragraph{Erasing Space Concept.}
We start with feature decomposition on the feature space to extract principal components.
We pass the forgetting data through the feature extractor of the original model ($h_{\psi}$) to obtain a feature matrix $\mathbf{Z}_f$~(see Figure~\ref{fig:main} (a)), which is then decomposed using SVD as follows:
\begin{equation}
\label{eq:svd}
    \mathbf{Z}_f=\mathbf{U}\cdot\mathbf{\Sigma}\cdot\mathbf{V}^\top,
\end{equation}
where $\mathbf{U}$ and $\mathbf{V}$ represent singular directions in the feature dimension and data length dimension, respectively, and $\mathbf{\Sigma}$ is the diagonal matrix containing singular values.
From Equation \ref{eq:svd}, we can get principal directions from the left singular vectors, \textit{i.e.}, $\textbf{U}=[u_1, u_2, \cdots, u_d] $, where $u_i \in \mathbb{R}^d$ is sorted based on the singular value, $\sigma_i $.
Based on our toy experiment, with the frozen backbone, our proposed \methodname{} projects features onto the subspace spanned by the pruned principal directions.
To obtain the pruned subspace, we eliminate the first $p$\% of principal directions from $\mathbf{U} $, as shown in Figure~\ref{fig:main} (b), where $p$ is a pruning hyperparameter.
Higher $p$ values correspond to larger knowledge deletion, and when $p=0 $, it indicates that the model remains identical to the original.
We can get the pruned principal directions $\mathbf{U}_P$ as:
\begin{equation}
\label{eq:pruned_U}
    \mathbf{U}_P=\mathbf{U}[k:],
\end{equation}
where $k$ is the number of pruned directions, determined by $p$ as $k=\frac{d}{100}\cdot p $, and it can vary depending on the embedding dimension. 
Using this selectively pruned principal directions, our unlearned feature extractor and the model by \methodname{} are defined as follows:
\vspace{-1mm}
\begin{align}
\label{eq:esc_model}
    \begin{split}
    h_{\psi_P}(x) = \mathbf{U}_P\cdot\mathbf{U}_P^\top\cdot h_{\psi}(x), \\
    f_{ESC}(x)=g_\phi \circ h_{\psi_P}(x).
    \end{split}
\end{align}
\vspace{-1mm}
From the extensive experiments, we demonstrate that even without any training process, this simple approach successfully deactivates the knowledge of forgetting data in both the output and feature spaces, \textit{i.e.}, \textit{\taskname{}} is fulfilled.

\subsection{Erasing Space Concept with Training}
\label{sec:esc_t}

While \methodname{} effectively deactivates the key features related to the forgetting data, removing entire vectors along the principal directions limits the flexibility of the removal process.
Hard thresholding, a na\"ive approach, can lead to over-forgetting; therefore, it is necessary to restrict the intensity of the removal selectively.
To this end, we propose \textbf{\methodname{}} with \textbf{T}raining (\textbf{\refinemethod{}}), which enhances \methodname{} through an additional lightweight training process that selectively blocks only the key elements in each principal direction related to the forgetting data.

In \refinemethod{}, we do not use the pruned principal directions in \methodname{}, which eliminates the entire vectors.
As shown in Figure~\ref{fig:main} (c-i), we start \refinemethod{} with creating a learnable refining mask $\mathbf{M}_0$, initialized with \textbf{1}, to obtain the optimized principal directions $\mathbf{U}_R$.

We then optimize the $\mathbf{M}_0$ by using our proposed forgetting loss (Figure~\ref{fig:main} (c-ii)).
Following the work of \cite{golatkar2020eternal, thudi2022unrolling, kim2023layer}, we design a \textit{Penalized Cross-Entropy Loss} for training the mask, which penalizes the mask when the model makes correct predictions.
This loss is defined as follows: 
\begin{table*}[t]
\centering
\centering
\resizebox{\textwidth}{!}{
\begin{tabular}{lcccccggg}
\toprule
\multirow{2}{*}{Method} & \multirow{2}{*}{$D_r$} & \multicolumn{6}{c}{CIFAR-10} \\ 
\cmidrule{3-9}
&& $D_f(\downarrow)$ & $D_r(\uparrow)$ & $D_{ft}(\downarrow)$ & $D_{rt}(\uparrow)$ & $HM(\uparrow)$ & $HM_{\text{t}}(\uparrow)$ & MIA\\
\midrule
Original & \multirow{2}{*}{-} & 98.42 & 98.29 & 85.90 & 86.57 & 3.11 & 24.25 & 57.68 \scriptsize$\pm$ 0.67 \\
Retrain && 0.00 & 96.93 & 0.00 & 86.02 & 98.44 & 92.48 & 50.06 \scriptsize$\pm$ 0.43 \\
\midrule
Finetune~\cite{warnecke2021machine} & \multirow{5}{*}{\ding{51}} & 6.89 \scriptsize$\pm$ 1.71 & 98.90 \scriptsize$\pm$ 0.14 & 5.53 \scriptsize$\pm$ 1.27 & 85.94 \scriptsize$\pm$ 0.17 & 95.91 \scriptsize$\pm$ 0.97 & 90.00 \scriptsize$\pm$ 0.62 & \underline{53.13 \scriptsize$\pm$ 0.67} \\
Fisher~\cite{golatkar2020eternal} && 0.74 \scriptsize$\pm$ 0.65 & 96.16 \scriptsize$\pm$ 0.25 & 0.63 \scriptsize$\pm$ 0.51 & 81.22 \scriptsize$\pm$ 0.79 & 97.69 \scriptsize$\pm$ 0.18 & 89.38 \scriptsize$\pm$ 0.28 & 54.60 \scriptsize$\pm$ 0.95 \\
SCRUB~\cite{kurmanji2023towards} && 10.39 \scriptsize$\pm$ 3.10 & 97.15 \scriptsize$\pm$ 0.45 & 4.50 \scriptsize$\pm$ 1.99 & 83.33 \scriptsize$\pm$ 0.38 & 93.23 \scriptsize$\pm$ 1.89 & 89.00 \scriptsize$\pm$ 1.09 & 55.50 \scriptsize$\pm$ 1.46 \\
$\ell_1$-sparse~\cite{liu2023model} && 00.00 \scriptsize$\pm$ 0.00 & 89.95 \scriptsize$\pm$ 0.15 & 00.00 \scriptsize$\pm$ 0.00 & 84.81 \scriptsize$\pm$ 2.00 & 94.71 \scriptsize$\pm$ 0.30 & 91.78 \scriptsize$\pm$ 3.92 & 55.76 \scriptsize$\pm$ 1.25 \\
SalUn~\cite{fan2024salun} && 00.00 \scriptsize$\pm$ 0.00 & 98.86 \scriptsize$\pm$ 0.34 & 00.00 \scriptsize$\pm$ 0.00 & 86.80 \scriptsize$\pm$ 0.70 & \textbf{99.43 \scriptsize$\pm$ 0.69} & \textbf{92.94 \scriptsize$\pm$ 1.38} & 56.42 \scriptsize$\pm$ 0.46 \\
\midrule
NG~\cite{thudi2022unrolling} & \multirow{6}{*}{\ding{55}} & 2.46 \scriptsize$\pm$ 0.30 & 90.27 \scriptsize$\pm$ 0.16 & 2.17 \scriptsize$\pm$ 0.15 & 75.83 \scriptsize$\pm$ 0.05 & 93.76 \scriptsize$\pm$ 0.06 & 85.44 \scriptsize$\pm$ 0.03 & 53.80 \scriptsize$\pm$ 1.68 \\
RL~\cite{golatkar2020eternal} && 13.25 \scriptsize$\pm$ 0.44 & 98.21 \scriptsize$\pm$ 0.04 & 7.93 \scriptsize$\pm$ 0.46 & 80.43 \scriptsize$\pm$ 0.03 & 92.12 \scriptsize$\pm$ 0.25 & 85.85 \scriptsize$\pm$ 0.19 & 57.10 \scriptsize$\pm$ 1.73 \\
BS~\cite{chen2023boundary} && 10.98 \scriptsize$\pm$ 0.64 & 95.89 \scriptsize$\pm$ 1.60 & 13.43 \scriptsize$\pm$ 2.71 & 84.41 \scriptsize$\pm$ 1.12 & 92.32 \scriptsize$\pm$ 0.42 & 85.43 \scriptsize$\pm$ 0.73 & 57.26 \scriptsize$\pm$ 0.38 \\
LAU~\cite{kim2023layer} && 0.15 \scriptsize$\pm$ 0.01 & 92.84 \scriptsize$\pm$ 0.42 & 0.07 \scriptsize$\pm$ 0.05 & 82.28 \scriptsize$\pm$ 0.43 & 96.22 \scriptsize$\pm$ 0.22 & 90.25 \scriptsize$\pm$ 0.28 & 54.92 \scriptsize$\pm$ 0.49 \\
\textbf{\methodname (Ours)} && 9.46 \scriptsize$\pm$ 0.06 & 96.52 \scriptsize$\pm$ 0.01 & 10.73 \scriptsize$\pm$ 0.12 & 85.59 \scriptsize$\pm$ 0.01 & 93.43 \scriptsize$\pm$ 0.03 & 87.39 \scriptsize$\pm$ 0.06 & \ty{\textbf{53.02 \scriptsize$\pm$ 0.53}} \\
\textbf{\refinemethod (Ours)} && 0.00 \scriptsize$\pm$ 0.00& 97.23 \scriptsize$\pm$ 0.19& 0.00 \scriptsize$\pm$ 0.00& 86.08 \scriptsize$\pm$ 0.27& \ty{\underline{98.60 \scriptsize$\pm$ 0.10}} & \ty{\underline{92.51\scriptsize$\pm$ 0.16}} & 56.72 \scriptsize$\pm$ 0.24 \\
\bottomrule
\vspace{0.5mm}
\end{tabular}

\begin{tabular}{ccccgg}
\toprule
\multicolumn{6}{c}{CIFAR-10$-KR$} \\ \cmidrule{1-6}
$D_f(\downarrow)$ & $D_r(\uparrow)$ & $D_{ft}(\downarrow)$ & $D_{rt}(\uparrow)$ & $HM(\uparrow)$ & $HM_{\text{t}}(\uparrow)$ \\
\midrule
98.40 & 97.81 & 86.40 & 85.96 & 3.15 & 23.48 \\ 
41.28 & 96.18 & 44.40 & 84.38 & 72.92 & 67.03 \\ 
\midrule
82.99 \scriptsize$\pm$ 2.03 & 98.60 \scriptsize$\pm$ 0.05 & 69.57 \scriptsize$\pm$ 3.33 & 85.35 \scriptsize$\pm$ 0.14 & 28.97 \scriptsize$\pm$ 2.98 & 44.80 \scriptsize$\pm$ 3.68 \\ 
96.00 \scriptsize$\pm$ 0.05 & 95.76 \scriptsize$\pm$ 1.16 & 83.33 \scriptsize$\pm$ 0.42 & 81.81 \scriptsize$\pm$ 0.10 & 7.68 \scriptsize$\pm$ 0.10 & 27.69 \scriptsize$\pm$ 0.57 \\ 
80.41 \scriptsize$\pm$ 4.91 & 96.62 \scriptsize$\pm$ 0.33 & 65.53 \scriptsize$\pm$ 1.99 & 81.85 \scriptsize$\pm$ 0.29 & 32.57 \scriptsize$\pm$ 6.96 & 48.51 \scriptsize$\pm$ 2.03 \\ 
62.03 \scriptsize$\pm$ 2.09 & 88.57 \scriptsize$\pm$ 0.49 & 61.12 \scriptsize$\pm$ 0.97 & 85.29 \scriptsize$\pm$ 0.61 & 53.15 \scriptsize$\pm$ 2.07 & 53.42 \scriptsize$\pm$ 0.80 \\ 
86.83 \scriptsize$\pm$ 3.06 & 98.12 \scriptsize$\pm$ 0.28 & 83.81 \scriptsize$\pm$ 0.33 & 91.22 \scriptsize$\pm$ 2.09 & 23.22 \scriptsize$\pm$ 4.72 & 27.49 \scriptsize$\pm$ 0.38 \\ 
\midrule
93.21 \scriptsize$\pm$ 1.37 & 99.14 \scriptsize$\pm$ 0.05 & 69.23 \scriptsize$\pm$ 1.14 & 83.02 \scriptsize$\pm$ 0.22 & 12.69 \scriptsize$\pm$ 2.42 & 44.89 \scriptsize$\pm$ 1.24 \\ 
98.05 \scriptsize$\pm$ 0.16 & 99.70 \scriptsize$\pm$ 0.03 & 72.97 \scriptsize$\pm$ 0.72 & 82.52 \scriptsize$\pm$ 0.42 & 3.83 \scriptsize$\pm$ 0.31 & 40.72 \scriptsize$\pm$ 0.86 \\ 
96.71 \scriptsize$\pm$ 1.02 & 96.93 \scriptsize$\pm$ 0.44 & 83.20 \scriptsize$\pm$ 1.84 & 85.40 \scriptsize$\pm$ 0.31 & 6.34 \scriptsize$\pm$ 1.88 & 28.03 \scriptsize$\pm$ 2.53 \\ 
97.86 \scriptsize$\pm$ 0.44 & 97.39 \scriptsize$\pm$ 0.39 & 85.57 \scriptsize$\pm$ 0.62 & 85.71 \scriptsize$\pm$ 0.30 & 4.19 \scriptsize$\pm$ 0.85 & 24.70 \scriptsize$\pm$ 0.91 \\ 
8.53 \scriptsize$\pm$ 0.58 & 97.45 \scriptsize$\pm$ 0.24 & 8.23 \scriptsize$\pm$ 0.55 & 86.56 \scriptsize$\pm$ 0.22 & \ty{\textbf{94.36 \scriptsize$\pm$ 0.27}} & \ty{\textbf{89.08 \scriptsize$\pm$ 0.16}} \\ 
29.47 \scriptsize$\pm$ 2.55 & 97.25 \scriptsize$\pm$ 0.47 & 25.07 \scriptsize$\pm$ 2.78 & 86.12 \scriptsize$\pm$ 0.40 & \underline{81.74 \scriptsize$\pm$ 1.56} & \underline{80.12 \scriptsize$\pm$ 1.44} \\ 
\bottomrule
\vspace{0.5mm}
\end{tabular}
}
\centering
\resizebox{\textwidth}{!}{
\begin{tabular}{lcccccggg}
\toprule
\multirow{2}{*}{Method} & \multirow{2}{*}{$D_r$} & \multicolumn{6}{c}{CIFAR-100} \\ \cmidrule{3-9}
&& $D_f(\downarrow)$ & $D_r(\uparrow)$ & $D_{ft}(\downarrow)$ & $D_{rt}(\uparrow)$ & $HM(\uparrow)$ & $HM_{\text{t}}(\uparrow)$ & MIA\\
\midrule
Original && 96.66 & 95.71 & 78.50 & 79.70 & 6.45 & 33.86 & 49.23 \scriptsize$\pm$ 1.42 \\
Retrain && 0.00 & 88.78 & 00.00 & 78.87 & 94.06 & 88.19 & 63.17 \scriptsize$\pm$ 2.87 \\
\midrule
Finetune~\cite{warnecke2021machine} & \multirow{5}{*}{\ding{51}} & 0.00 \scriptsize$\pm$ 0.00 & 89.98 \scriptsize$\pm$ 0.42 & 0.00 \scriptsize$\pm$ 0.00 & 74.18 \scriptsize$\pm$ 0.25 & 94.72 \scriptsize$\pm$ 0.87 & 85.18 \scriptsize$\pm$ 0.50 & 69.12 \scriptsize$\pm$ 0.97 \\
Fisher~\cite{golatkar2020eternal} && 0.00 \scriptsize$\pm$ 0.00 & 67.46 \scriptsize$\pm$ 0.39 & 00.00 \scriptsize$\pm$ 0.00 & 54.59 \scriptsize$\pm$ 0.59 & 80.57 \scriptsize$\pm$ 0.28 & 70.63 \scriptsize$\pm$ 0.49 & 55.28 \scriptsize$\pm$ 1.40 \\
SCRUB~\cite{kurmanji2023towards} && 19.15 \scriptsize$\pm$ 0.05 & 86.98 \scriptsize$\pm$ 1.69 & 14.14 \scriptsize$\pm$ 0.93 & 75.82 \scriptsize$\pm$ 0.62 & 83.80 \scriptsize$\pm$ 0.79 & 80.53 \scriptsize$\pm$ 0.58 & 56.33 \scriptsize$\pm$ 0.67 \\
$\ell_1$-sparse~\cite{liu2023model} && 0.00 \scriptsize$\pm$ 0.00 & 57.34 \scriptsize$\pm$ 0.14 & 0.00 \scriptsize$\pm$ 0.00 & 50.59 \scriptsize$\pm$ 0.41 & 72.89 \scriptsize$\pm$ 0.28 & 67.79 \scriptsize$\pm$ 0.81 & 56.42 \scriptsize$\pm$ 0.56 \\
SalUn~\cite{fan2024salun} && 0.76 \scriptsize$\pm$ 0.00 & 96.64 \scriptsize$\pm$ 0.01 & 2.33 \scriptsize$\pm$ 0.02 & 67.73 \scriptsize$\pm$ 0.88 & \textbf{97.93 \scriptsize$\pm$ 0.00} & 79.99 \scriptsize$\pm$ 0.61 & 56.12 \scriptsize$\pm$ 0.72 \\
\midrule
NG~\cite{thudi2022unrolling} & \multirow{6}{*}{\ding{55}} & 13.24 \scriptsize$\pm$ 0.61 & 84.50 \scriptsize$\pm$ 0.32 & 10.87 \scriptsize$\pm$ 0.42 & 71.64 \scriptsize$\pm$ 0.08 & 85.62 \scriptsize$\pm$ 0.64 & 79.43 \scriptsize$\pm$ 0.15 & 54.97 \scriptsize$\pm$ 0.59 \\
RL~\cite{golatkar2020eternal} && 27.45 \scriptsize$\pm$ 0.53 & 84.74 \scriptsize$\pm$ 0.43 & 19.33 \scriptsize$\pm$ 0.35 & 70.38 \scriptsize$\pm$ 0.21 & 78.17 \scriptsize$\pm$ 0.24 & 75.17 \scriptsize$\pm$ 0.10 & 53.23 \scriptsize$\pm$ 1.46 \\
BS~\cite{chen2023boundary} && 9.93 \scriptsize$\pm$ 0.79 & 61.16 \scriptsize$\pm$ 1.14 & 7.23 \scriptsize$\pm$ 0.74 & 50.53 \scriptsize$\pm$ 1.03 & 72.85 \scriptsize$\pm$ 0.56 & 65.42 \scriptsize$\pm$ 0.69 & 54.30 \scriptsize$\pm$ 1.97 \\
LAU~\cite{kim2023layer} && 0.00 \scriptsize$\pm$ 0.00 & 44.14 \scriptsize$\pm$ 6.55 & 0.00 \scriptsize$\pm$ 0.00 & 36.84 \scriptsize$\pm$ 5.10 & 61.25 \scriptsize$\pm$ 6.47 & 53.84 \scriptsize$\pm$ 5.56 & \ty{\textbf{60.53 \scriptsize$\pm$ 0.85}} \\
\textbf{\methodname (Ours)} && 0.80 \scriptsize$\pm$ 0.14 & 89.46 \scriptsize$\pm$ 0.19 & 0.50 \scriptsize$\pm$ 0.1 & 74.21 \scriptsize$\pm$ 0.17 & 94.08 \scriptsize$\pm$ 0.17 & 85.01 \scriptsize$\pm$ 0.15 & 58.37 \scriptsize$\pm$ 0.38 \\
\textbf{\refinemethod (Ours)} && 0.00 \scriptsize$\pm$ 0.00 & 91.20 \scriptsize$\pm$ 0.55 & 0.09 \scriptsize$\pm$ 0.01 & 76.73 \scriptsize$\pm$ 0.84 & \ty{\underline{95.40 \scriptsize$\pm$ 0.30}} & \ty{\textbf{86.80 \scriptsize$\pm$ 0.53}} & \underline{58.93 \scriptsize$\pm$ 1.27} \\
\bottomrule
\vspace{0.5mm}
\end{tabular}

\begin{tabular}{ccccgg}
\toprule
\multicolumn{6}{c}{CIFAR-100$-KR$} \\ \cmidrule{1-6}
$D_f(\downarrow)$ & $D_r(\uparrow)$ & $D_{ft}(\downarrow)$ & $D_{rt}(\uparrow)$ & $HM(\uparrow)$ & $HM_{\text{t}}(\uparrow)$ \\
\midrule
83.06 & 83.27 & 69.40 & 70.54 & 28.15 & 42.68 \\ 
54.66 & 82.03 & 87.80 & 73.10 & 58.40 & 53.51 \\ 
\midrule
40.68 \scriptsize$\pm$ 1.60 & 78.21 \scriptsize$\pm$ 0.27 & 38.25 \scriptsize$\pm$ 0.20 & 72.88 \scriptsize$\pm$ 11.16 & 67.47 \scriptsize$\pm$ 1.07 & 66.85 \scriptsize$\pm$ 4.38 \\ 
81.46 \scriptsize$\pm$ 2.57 & 84.05 \scriptsize$\pm$ 1.63 & 60.96 \scriptsize$\pm$ 1.60 & 70.24 \scriptsize$\pm$ 1.35 & 30.38 \scriptsize$\pm$ 3.57 & 50.18 \scriptsize$\pm$ 1.67 \\ 
74.85 \scriptsize$\pm$ 0.46 & 79.25 \scriptsize$\pm$ 0.22 & 55.27 \scriptsize$\pm$ 0.21 & 65.57 \scriptsize$\pm$ 0.46 & 38.19 \scriptsize$\pm$ 0.51 & 53.18 \scriptsize$\pm$ 0.19 \\ 
60.56 \scriptsize$\pm$ 0.03 & 65.78 \scriptsize$\pm$ 0.32 & 54.63 \scriptsize$\pm$ 0.09 & 58.53 \scriptsize$\pm$ 0.43 & 49.31 \scriptsize$\pm$ 0.63 & 51.11 \scriptsize$\pm$ 0.86 \\ 
86.68 \scriptsize$\pm$ 1.95 & 96.61 \scriptsize$\pm$ 0.32 & 66.86 \scriptsize$\pm$ 1.77 & 71.86 \scriptsize$\pm$ 0.89 & 23.41 \scriptsize$\pm$ 2.98 & 45.36 \scriptsize$\pm$ 1.80 \\ 
\midrule
52.39 \scriptsize$\pm$ 0.33 & 70.68 \scriptsize$\pm$ 0.15 & 44.24 \scriptsize$\pm$ 0.10 & 62.19 \scriptsize$\pm$ 0.21 & 56.90 \scriptsize$\pm$ 0.19 & 58.80 \scriptsize$\pm$ 0.12 \\ 
46.48 \scriptsize$\pm$ 0.31 & 70.76 \scriptsize$\pm$ 0.22 & 38.88 \scriptsize$\pm$ 0.50 & 61.48 \scriptsize$\pm$ 0.27 & 60.95 \scriptsize$\pm$ 0.14 & 61.30 \scriptsize$\pm$ 0.38 \\ 
55.98 \scriptsize$\pm$ 1.56 & 72.63 \scriptsize$\pm$ 0.61 & 46.63 \scriptsize$\pm$ 1.60 & 62.42 \scriptsize$\pm$ 0.54 & 54.81 \scriptsize$\pm$ 1.32 & 57.54 \scriptsize$\pm$ 0.99 \\ 
81.81 \scriptsize$\pm$ 0.58 & 83.14 \scriptsize$\pm$ 0.12 & 68.43 \scriptsize$\pm$ 1.80 & 70.68 \scriptsize$\pm$ 0.62 & 29.85 \scriptsize$\pm$ 0.78 & 43.64 \scriptsize$\pm$ 1.71 \\ 
0.27 \scriptsize$\pm$ 0.01 & 76.55 \scriptsize$\pm$ 0.45 & 0.59 \scriptsize$\pm$ 0.02 & 65.74 \scriptsize$\pm$ 0.36 & \ty{\textbf{86.62 \scriptsize$\pm$ 0.28}} & \underline{79.14 \scriptsize$\pm$ 0.21} \\ 
8.59 \scriptsize$\pm$ 0.59 & 80.46 \scriptsize$\pm$ 0.49 & 8.05 \scriptsize$\pm$ 0.09 & 71.86 \scriptsize$\pm$ 0.89 & \underline{85.59 \scriptsize$\pm$ 0.24} & \ty{\textbf{80.67 \scriptsize$\pm$ 0.79}} \\ 
\bottomrule
\vspace{0.5mm}
\end{tabular}
}
\centering
\resizebox{\textwidth}{!}{
\begin{tabular}{lcccccggg}
\toprule
\multirow{2}{*}{Method} & \multirow{2}{*}{$D_r$} & \multicolumn{6}{c}{Tiny-ImageNet} \\ \cmidrule{3-9}
&& $D_f(\downarrow)$ & $D_r(\uparrow)$ & $D_{ft}(\downarrow)$ & $D_{rt}(\uparrow)$ & $HM(\uparrow)$ & $HM_{\text{t}}(\uparrow)$ & MIA\\
\midrule
Original && 96.72 & 96.47 & 90.30 & 90.24 & 6.34 & 17.52 & 55.80 \scriptsize$\pm$ 0.62 \\
Retrain && 0.00 & 96.93 & 0.00 & 86.02 & 98.44 & 92.48 & 51.33 \scriptsize$\pm$ 0.59 \\
\midrule
Finetune~\cite{warnecke2021machine} & \ding{51} & 20.58 \scriptsize$\pm$ 0.61 & 96.95 \scriptsize$\pm$ 0.06 & 18.77 \scriptsize$\pm$ 0.61 & 84.55 \scriptsize$\pm$ 0.58 & 87.31 \scriptsize$\pm$ 0.11 & 82.86 \scriptsize$\pm$ 1.15 & 59.40 \scriptsize$\pm$ 2.27 \\
\midrule
NG~\cite{thudi2022unrolling} & \multirow{6}{*}{\ding{55}} & 0.00 \scriptsize$\pm$ 0.00 & 0.56 \scriptsize$\pm$ 0.00 & 00.00 \scriptsize$\pm$ 0.00 & 0.56 \scriptsize$\pm$ 0.00 & 1.11 \scriptsize$\pm$ 0.00 & 1.11 \scriptsize$\pm$ 0.00 & 48.57 \scriptsize$\pm$ 1.86 \\
RL~\cite{golatkar2020eternal} && 5.13 \scriptsize$\pm$ 0.84 & 5.76 \scriptsize$\pm$ 0.52 & 5.57 \scriptsize$\pm$ 0.76 & 5.71 \scriptsize$\pm$ 0.61 & 10.86 \scriptsize$\pm$ 0.93 & 10.76 \scriptsize$\pm$ 1.08 & \underline{52.90 \scriptsize$\pm$ 1.14} \\
BS~\cite{chen2023boundary} && 11.25 \scriptsize$\pm$ 0.41 & 14.52 \scriptsize$\pm$ 1.21 & 10.03 \scriptsize$\pm$ 0.23 & 14.54 \scriptsize$\pm$ 1.09 & 24.96 \scriptsize$\pm$ 1.78 & 25.03 \scriptsize$\pm$ 1.61 & 60.07 \scriptsize$\pm$ 0.47 \\
LAU~\cite{kim2023layer} && 0.00 \scriptsize$\pm$ 0.00 & 95.62 \scriptsize$\pm$ 0.01 & 0.00 \scriptsize$\pm$ 0.00 & 89.76 \scriptsize$\pm$ 0.03 & 97.76 \scriptsize$\pm$ 0.00 & 94.60 \scriptsize$\pm$ 0.02 & 49.03 \scriptsize$\pm$ 1.84 \\
\textbf{\methodname (Ours)} && 0.10 \scriptsize$\pm$ 0.02 & 96.36 \scriptsize$\pm$ 0.05 & 0.03 \scriptsize$\pm$ 0.06 & 90.57 \scriptsize$\pm$ 0.05 & \underline{98.10 \scriptsize$\pm$ 0.02} & \underline{95.03 \scriptsize$\pm$ 0.05} & \ty{\textbf{50.20 \scriptsize$\pm$ 3.06}} \\
\textbf{\refinemethod (Ours)} && 0.00 \scriptsize$\pm$ 0.00 & 96.49 \scriptsize$\pm$ 0.23 & 0.00 \scriptsize$\pm$ 0.00 & 90.57 \scriptsize$\pm$ 0.12 & \ty{\textbf{98.21 \scriptsize$\pm$ 0.12}} & \ty{\textbf{95.05 \scriptsize$\pm$ 0.07}} & 56.73 \scriptsize$\pm$ 1.55 \\
\bottomrule
\end{tabular}

\begin{tabular}{ccccgg}
\toprule
\multicolumn{6}{c}{Tiny-ImageNet$-KR$} \\ \cmidrule{1-6}
$D_f(\downarrow)$ & $D_r(\uparrow)$ & $D_{ft}(\downarrow)$ & $D_{rt}(\uparrow)$ & $HM(\uparrow)$ & $HM_{\text{t}}(\uparrow)$ \\
\midrule
95.55 & 94.80 & 90.50 & 88.57 & 8.50 & 17.16 \\ 
66.56 & 92.71 & 63.30 & 82.79 & 49.15 & 50.86 \\ 
\midrule
64.61 \scriptsize$\pm$ 0.83 & 93.40 \scriptsize$\pm$ 0.89 & 62.73 \scriptsize$\pm$ 0.38 & 81.84 \scriptsize$\pm$ 0.57 & 51.33 \scriptsize$\pm$ 0.89 & 51.21 \scriptsize$\pm$ 0.43 \\ 
\midrule
1.25 \scriptsize$\pm$ 0.83 & 0.96 \scriptsize$\pm$ 0.05 & 1.37 \scriptsize$\pm$ 0.70 & 0.89 \scriptsize$\pm$ 0.11 & 1.90 \scriptsize$\pm$ 0.10 & 1.76 \scriptsize$\pm$ 0.21 \\ 
8.44 \scriptsize$\pm$ 1.55 & 7.49 \scriptsize$\pm$ 1.40 & 8.23 \scriptsize$\pm$ 0.45 & 7.23 \scriptsize$\pm$ 1.73 & 13.85 \scriptsize$\pm$ 2.43 & 13.40 \scriptsize$\pm$ 3.02 \\ 
40.85 \scriptsize$\pm$ 2.63 & 39.67 \scriptsize$\pm$ 1.90 & 39.63 \scriptsize$\pm$ 2.83 & 39.37 \scriptsize$\pm$ 1.36 & 47.49 \scriptsize$\pm$ 1.30 & 47.66 \scriptsize$\pm$ 1.63 \\ 
95.52 \scriptsize$\pm$ 0.20 & 94.77 \scriptsize$\pm$ 0.05 & 89.33 \scriptsize$\pm$ 0.85 & 88.67 \scriptsize$\pm$ 0.04 & 8.56 \scriptsize$\pm$ 0.36 & 19.04 \scriptsize$\pm$ 1.36 \\ 
0.27 \scriptsize$\pm$ 0.11 & 94.64 \scriptsize$\pm$ 0.10 & 0.20 \scriptsize$\pm$ 0.17 & 88.88 \scriptsize$\pm$ 0.02 & \ty{\textbf{97.19 \scriptsize$\pm$ 0.06}} & \ty{\textbf{94.02 \scriptsize$\pm$ 0.04}} \\ 
1.36 \scriptsize$\pm$ 0.07 & 94.53 \scriptsize$\pm$ 0.28 & 0.87 \scriptsize$\pm$ 0.04 & 88.81 \scriptsize$\pm$ 0.16 & \underline{96.54 \scriptsize$\pm$ 0.17} & \underline{93.69 \scriptsize$\pm$ 0.08} \\ 
\bottomrule
\end{tabular}
}
\caption{Accuracy, MIA, and KR performance in the KD setting are evaluated using CIFAR-10 with AllCNN, CIFAR-100 with ResNet-18, and Tiny-ImageNet with ViT. $D_r$ indicates which methods use the remaining data during the unlearning process. The table presents the mean and standard deviation (mean \scriptsize$\pm$ std\small) across three trials. The best value is highlighted in bold, while the second-best is underlined. Additionally, the best result among methods that do not use $D_r$ is colored in blue.}
\label{tab:main}
\end{table*}
\begin{equation}
\label{eq:pce_loss}
  \mathcal{L}_{PCE}=
  \begin{cases}
    -\mathcal{L}_{CE}(f_{\theta}(x),y), & \text{if $\hat{y}=y $},\\
    0, & \text{otherwise},
  \end{cases}
\end{equation}
where $\hat{y}$ is the predicted class from the model $f_\theta $.
We update only the mask to minimize $\mathcal{L}_{PCE}$ until either all predictions of the unlearned model are incorrect for the given forgetting data, or a specified number of steps has been reached.
Additionally, at each training step $t$, we constrain the values of all elements in the mask $\mathbf{M}_t$ between 0 and 1 to maintain stable feature space.
When the learning process is done, we apply a threshold $\tau$ to modify the mask $\mathbf{M}_T$ to be binary, \textit{i.e.}, containing only 0s and 1s, resulting in the final binary mask $\mathbf{M}_R$ (Figure~\ref{fig:main} (c-iii)).
This last step can completely remove the influence of important elements for forgetting knowledge while preserving unimportant elements.
Based on $\mathbf{M}_R$, we modify Equation \ref{eq:pruned_U} as follows:
\vspace{-0.5mm}
\begin{equation}
\label{eq:refine_U}
    \mathbf{U}_R=\mathbf{U} \odot \mathbf{M}_R,
\end{equation}
where $\mathbf{U}_R$ is the trained principal directions and $\odot$ is the Hadamard product.
We then finally get \refinemethod{} as follows:
\begin{align}
\vspace{-3mm}
\label{eq:esct_model}
    \begin{split}
    &h_{\psi_R}(x) = \mathbf{U}_R\cdot\mathbf{U}_R^\top\cdot h_{\psi}(x), \\
    &f_{ESC-T}(x)=g_\phi \circ h_{\psi_R}(x).
    \end{split}
\end{align}
We provide a detailed description of Algorithm 1 for \refinemethod{} in the supplementary~\ref{supp:algorithm}.
Through this simple yet efficient learning step, \refinemethod{} enhances the removal intensity without degrading the model's performance. 
In Figure~\ref{fig:main} (d), we describe the inference process of our methods.

\section{Experiments}
\label{sec:experiments}

\subsection{Experimental Setup}
\label{sec:experimental_setup}

\paragraph{Datasets, Models, and Setting.}
We conducted unlearning on the CIFAR-10~\cite{krizhevsky2009learning}, CIFAR-100~\cite{krizhevsky2009learning}, Tiny-ImageNet~\cite{le2015tiny}, and Lacuna-10~\cite{golatkar2020eternal,kurmanji2023towards} datasets.
In the case of KD, we removed 10\% of the total number of classes, while for random data forgetting, we unlearned 10\% of the data.
Experiments were conducted on various models, including All-CNN~\cite{springenberg2014striving}, ResNet-18~\cite{he2016deep} pre-trained on ImageNet-1K~\cite{russakovsky2015imagenet}, and Vision Transformer~(ViT)~\cite{dosovitskiy2021image} pre-trained on ImageNet-21K~\cite{ridnik2021imagenet21k}.
We conducted all experiments using a single NVIDIA A5000 GPU.
Further details on datasets and setups are available in the supplementary~\ref{supp:add_details}.

\vspace{-3mm}
\paragraph{Baselines.} We compared our methods with various approximate unlearning methods.
\textbf{Original} is a complete model trained using $\mathcal{D}$ from scratch or pre-trained weights.
\textbf{Retrain} refers to a model trained from scratch using only $\mathcal{D}_r$. However, for ResNet-18 and ViT, we initialize from a pre-trained model, so it is not strictly the same with retrain-from-scratch.
We focus on real-world scenarios and primarily compared our approach with existing MU methods, including \textbf{NG}~\cite{thudi2022unrolling}, \textbf{RL}~\cite{golatkar2020eternal}, \textbf{BS}~\cite{chen2023boundary}, and \textbf{LAU}~\cite{kim2023layer}, all of which use only the forgetting data.
For more extensive experiments, we also included \textbf{Finetune}~\cite{warnecke2021machine}, \textbf{Fisher}~\cite{golatkar2020eternal}, \textbf{SCRUB}~\cite{kurmanji2023towards}, \textbf{$\ell_1$-sparse}~\cite{liu2023model}, and \textbf{SalUn}~\cite{fan2024salun}, although all of these methods require the use of the entire remaining data.


\vspace{-3mm}
\paragraph{Evaluation Metrics.}
In our experiments, we divided the datasets into remaining training data $\mathcal{D}_r$, forgetting training data $\mathcal{D}_f$, remaining testing data $\mathcal{D}_{rt}$, and forgetting testing data $\mathcal{D}_{ft}$. 
We calculated accuracy for each dataset segment both in post-unlearning and the proposed $\metricname{}$.
Furthermore, similar to~\cite{kim2023layer}, we utilized the Harmonic Mean ($HM$) for comprehensive evaluation.
To assess the privacy guarantee, we also constructed MIA, widely used in MU.

\subsection{Experimental Results}

\begin{figure}[!t]
    \centering
    \includegraphics[width=\columnwidth]{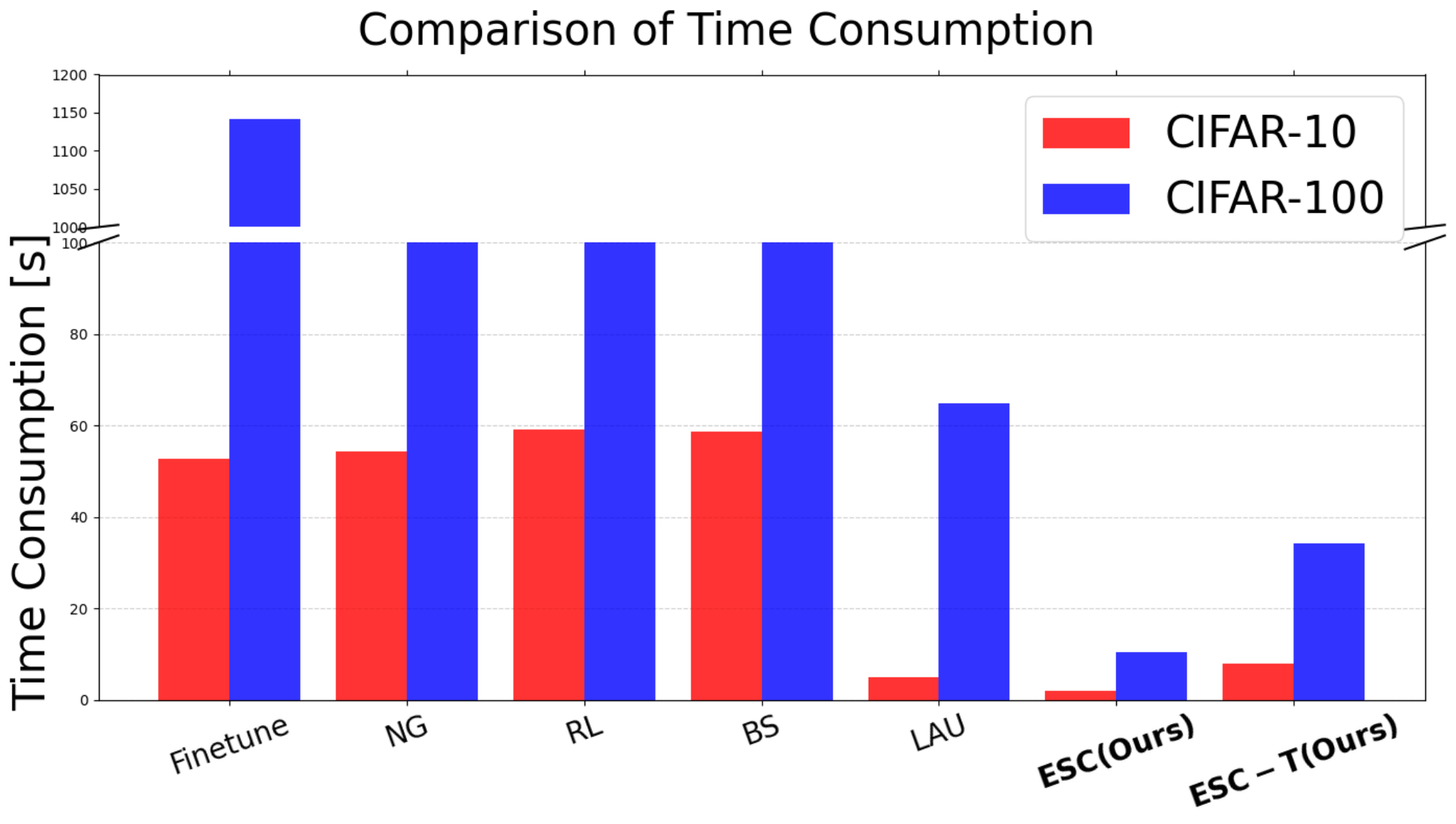}
    \vspace{-7mm}
    \caption{Comparison of time consumption.
    Our approaches are the most efficient method compared with others.
    Additional results reported in Supplementary~\ref{supp:add_exp}.}
    \vspace{-4mm}
    \label{fig:time} 
\end{figure}

We present comprehensive experimental results of \taskname{} across multiple datasets and models, compared with various MU methods, even though some of these methods essentially required the entire datasets, as shown in Table~\ref{tab:main}.
The results indicate that our methods achieved the best utility-focused performance in all cases compared to methods that did not use $D_r$.
Even when compared to the method using $D_r$, our approach showed comparable results.
The results also demonstrated that \refinemethod{} achieved a better balance between forgetting and preservation compared to \methodname{} by selectively removing elements in the principal directions through a simple training process. 
In privacy-focused MIA, our approach also demonstrated performance similar to that of other MU methods overall.
We additionally reported the ZRF~\cite{chundawat2023can} metric, an additional privacy-related evaluation, in the supplementary~\ref{supp:add_exp}.

In the \metricname{} setting, all baseline methods, even those using $D_r$, suffered from substantial knowledge retention in the features.
$\ell_1$-sparse showed comparable results among the MU methods across all experimental settings, as it directly prunes model parameters for unlearning. 
However, as demonstrated in our analysis in Figure~\ref{fig:feasibility},  the original model features are highly entangled across different classes, making pruning insufficient to effectively eliminate the forgetting knowledge.
In contrast, our methods significantly outperformed other baselines, and we effectively eliminated the forgetting knowledge in the feature space while preserving the remaining knowledge.


In Figure~\ref{fig:time}, we present the time consumption during the \taskname{}.
\methodname{} achieved the smallest time consumption because it required only a single forward process.
\refinemethod{} was also efficient, as it involved training only the small parameters of the learnable mask with a high learning rate.
Specifically, \methodname{} is 9.17 times faster than the previously fastest method, LAU, across all settings.
We highlight again that our methods only use forgetting data, so that it is a more practical solution in the real world.

\vspace{-1mm}
\subsection{Ablation Study}

\begin{table}[!t]
    \centering
    \resizebox{0.8\columnwidth}{!}{
    \begin{tabular}{ccc}
    \toprule
    \multirow{2}{*}{\(p\) (\%)} & \multicolumn{2}{c}{CIFAR-100} \\ \cmidrule{2-3}
    &\(HM(\uparrow)\)&\(HM_{\text{t}}(\uparrow)\)\\
    \midrule
        \textbf{1.7} & \textbf{99.01} & \textbf{96.45} \\
        5 & 98.58 & 96.00 \\
        {10}  & {98.30}  & {95.53} \\
        50  & 95.75  & 92.99 \\
    \bottomrule
    \end{tabular}
    \begin{tabular}{cc}
    \toprule
    \multicolumn{2}{c}{CIFAR-100$-KR$} \\ \cmidrule{1-2}
   \(HM(\uparrow)\)&\(HM_{\text{t}}(\uparrow)\)\\
    \midrule
        %
       \textbf{98.30} & \textbf{95.70} \\ 
        97.30 & 94.69 \\ 
        {95.19}  & {92.61} \\ 
        76.66  & 74.05 \\ 
    \bottomrule
    \end{tabular}
}
\caption{Impact of pruning hyperparameter \(p\) in \methodname{}. The results show that \methodname{} is robust to pruning hyperparameter \(p\).}
\vspace{-2mm}
\label{tab:p}
\end{table}

\paragraph{$p$ Ablation.}
We conducted an ablation study on the pruning hyperparameter $p$ in \methodname{}.
As $p$ approaches 0\%, more knowledge about the forgetting data is eliminated, which can affect the remaining accuracy. 
In contrast, as the $p$ approaches 100\%, it eliminates more knowledge about the forgetting data and can influence the remaining accuracy.
As shown in Table~\ref{tab:p}, \methodname{} performed robustly across different values of $p$, and additional results are provided in the supplementary material~\ref{supp:add_abl}.


\begin{table}[tb]
\centering
\resizebox{\columnwidth}{!}{
\begin{tabular}{cccccc}
\toprule
\multirow{2}{*}{Method } & \multirow{2}{*}{\(\mathcal{L}_{PCE}\)}& \multirow{2}{*}{Clip} & \multirow{2}{*}{\(\tau\)}& \multicolumn{2}{c}{CIFAR-100} \\ \cmidrule{5-6}
   & & & & \(HM(\uparrow)\) & \(HM_{\text{t}}(\uparrow)\) \\
    \midrule
\methodname{}   &            &           &           & 99.01 & 96.45\\ 
\hdashline
                & \checkmark  &           &           & 99.05 & 96.41 \\
                & \checkmark  & \checkmark&           & 99.04 & 96.46 \\
                & \checkmark  &           & \checkmark& 90.33  & 89.60 \\
\refinemethod{} & \checkmark  & \checkmark& \checkmark& \textbf{99.05} & \textbf{96.47}\\ 
\bottomrule
\end{tabular}

\begin{tabular}{cccc}
\toprule
\multicolumn{2}{c}{CIFAR-100\(-KR\)} \\ \cmidrule{1-2}
    \(HM(\uparrow)\) & \(HM_{\text{t}}(\uparrow)\) \\
    \midrule
    98.30 & 95.70 \\
    \hdashline
    92.03 & 89.59 \\
    94.70 & 91.59\\
    97.48 & 94.22 \\
    \textbf{97.56} & \textbf{95.11}\\
\bottomrule
\end{tabular}
}
\caption{Ablation table to figure out the effectiveness of our training method in \refinemethod{}. The results clearly show the effectiveness of each components in \refinemethod{}.}
\vspace{-4mm}
\label{tab:clipping}
\end{table}

\vspace{-2mm}
\paragraph{Ablation Study for \refinemethod{}.}
To assess the effectiveness of each component of \refinemethod{}, we present the ablation study results in Table~\ref{tab:clipping}.
Mask learning using $\mathcal{L}_{PCE}$ enhanced the \taskname{} performance; however, this improvement alone was insufficient.
Introducing clipping to restrict mask values within the range of 0 to 1 led to an improvement in \metricname{} performance.
Applying a threshold effectively prevented recovery but resulted in a degradation of accuracy. 
Notably, when all components were used together, \refinemethod{}, the overall performance was the best.
We additionally conducted analysis about the clipping method in the supplementary~\ref{supp:add_abl}.

\begin{table}[!t]
    \centering
    \resizebox{0.8\columnwidth}{!}{
    \begin{tabular}{lcc}
    \toprule
    \multirow{2}{*}{$\tau$} & \multicolumn{2}{c}{CIFAR-100} \\ \cmidrule{2-3}
    & $HM(\uparrow)$ & $HM_{\text{t}}(\uparrow)$ \\
    \midrule
    0.6 & 98.60  & 95.95 \\
    0.7 &  \underline{99.05}  & \underline{96.43} \\
    0.8 & \textbf{99.06} & \textbf{96.49}\\
    0.9 & 98.26  & 95.44 \\
    \bottomrule
    \end{tabular}
    
    \begin{tabular}{cc}
    \toprule
    \multicolumn{2}{c}{CIFAR-100$-KR$} \\ \cmidrule{1-2}
    $HM(\uparrow)$ & $HM_{\text{t}}(\uparrow)$ \\
    \midrule
    \underline{97.38} & \textbf{95.13} \\ 
    \textbf{97.49} & \textbf{95.13} \\ 
    96.86 & 94.09\\ 
    58.27 & 58.79 \\ 
    \bottomrule
    \end{tabular}
}
\vspace{-2mm}
\caption{Changes in unlearning performance with ViT based on hyper-parameter $\tau$ of \refinemethod{}.}
\vspace{-2mm}
\label{tab:threshold}
\end{table}

\vspace{-1mm}
\paragraph{Sensitivity Analysis of Threshold \texorpdfstring{$\mathcal{T}$}{T}.}
In Table~\ref{tab:threshold}, we present the robustness of our approach to changes in the threshold parameter $\tau$.
These results demonstrated that our proposed method is not sensitive to variations in $\tau$. 
In our experiments, we used a value of $\tau$ between 0.7 and 0.8, which was empirically found.

\subsection{Additional Experiments}

We additionally conducted a comprehensive analysis to evaluate generalizability. 
Due to space limitations, we only report results for random data forgetting and partial visualizations. 
In the supplementary~\ref{supp:add_exp}, we provide further scenarios, such as incremental KD, along with additional combinations of datasets and models.

\begin{table}[t]
\centering
\resizebox{\columnwidth}{!}{
\begin{tabular}{lcccccgc}
\toprule
\multirow{2}{*}{Method} & \multirow{2}{*}{$D_r$} & \multicolumn{6}{c}{CIFAR-10 Random Data Forgetting (10\%)} \\ \cmidrule{3-8}
&& $D_f(\downarrow)$ & $D_r(\uparrow)$ & $D_{t}(\downarrow)$ & MIA & Avg. Gap & RTE(s) \\
\midrule
Retrain && 95.58 & 100.00 & 95.23 & 74.64 & - & 2681.72 \\
\midrule
Finetune~\cite{warnecke2021machine} & \multirow{5}{*}{\ding{55}} & 99.85 & 100.00 & 95.08 & 87.73 & 4.38 & 132.02 \\
RL~\cite{golatkar2020eternal} && 94.52& 99.97 & 93.66 & 27.99 & 12.33 & 2.57 \\
SCRUB~\cite{kurmanji2023towards} && 99.99 & 100.00 & 95.41 & 86.41 & 4.09 & 160.39 \\
BadT~\cite{chundawat2023can} && 100.00 & 100.00 & 95.27 & 60.33 & 4.69 & 182.78 \\
SalUn~\cite{fan2024salun} && 100.00 & 99.99 & 95.18 & 63.61 & 3.88 & 150.51 \\
\midrule
NG~\cite{thudi2022unrolling} & \multirow{3}{*}{\ding{51}} & 96.57 & 96.59 & 89.67 & 79.26 & 3.65 & 26.79 \\
\textbf{\methodname (Ours)} && 100.00 & 100.00 & 95.07 & 73.43 & \ty{\textbf{1.45}} & 2.24 \\
\textbf{\refinemethod (Ours)} && 99.86 & 97.78 & 92.73 & 76.74 & \underline{2.77} & 29.47 \\
\bottomrule
\end{tabular}
}
\vspace{-1mm}
\caption{Accuracy, MIA, and Run Time Efficiency (RTE) performance using ResNet-18 on CIFAR-10. The table presents the average of across three trials. The best value is in bold, the second-best is underlined, and the best without $D_r$ is in blue.}
\vspace{-4mm}
\label{tab:random}
\end{table}

\paragraph{Random Data Forgetting.}
To demonstrate the generalizability of our methods, we also conducted a random data forgetting scenario, aimed at removing the influence of specific data. 
As shown in Table~\ref{tab:random}, our methods demonstrated superior performance compared to other approaches (see the average performance gap). 
Notably, our methods achieved this using only the forgetting data, without compromising the overall performance of the unlearned model.

\begin{figure}[!t]
    \centering
    \includegraphics[width=\columnwidth]{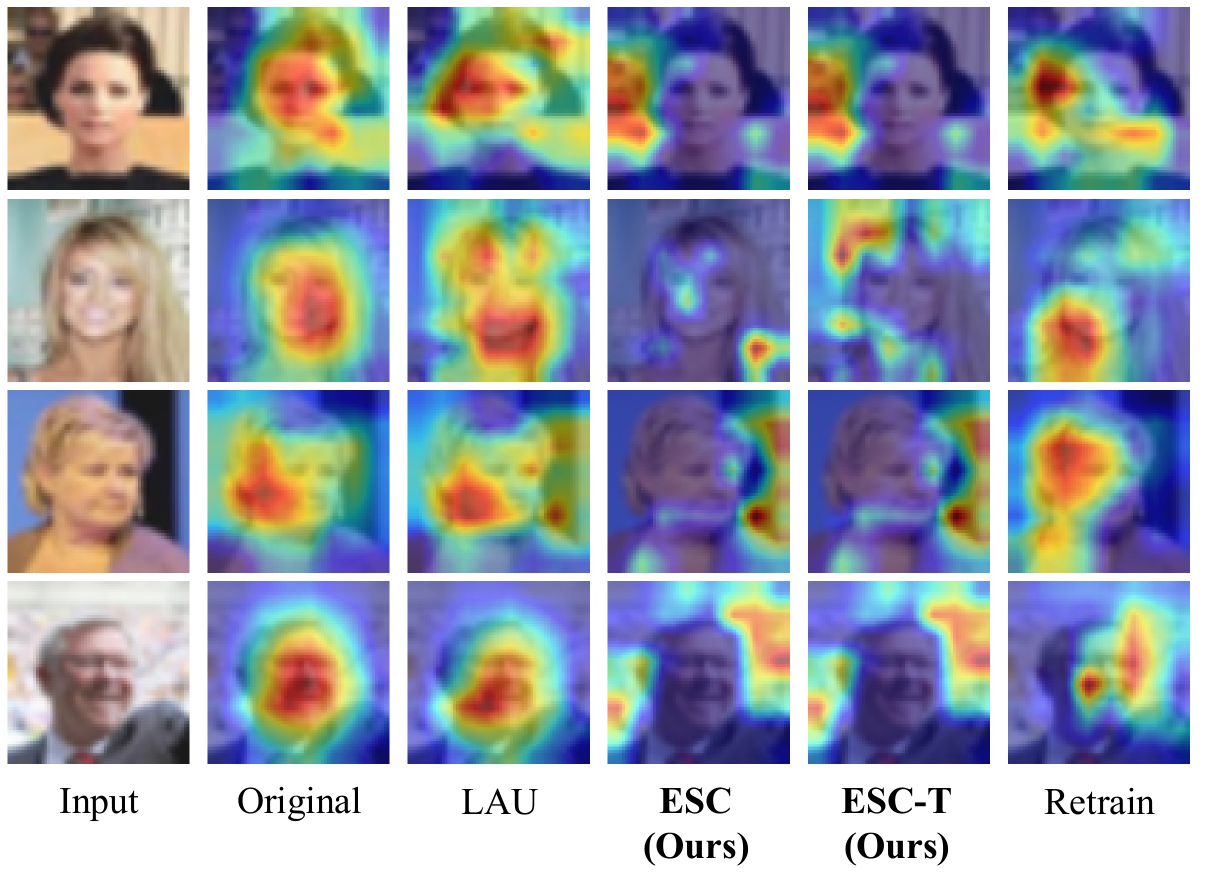}
    \vspace{-6mm}
    \caption{The activation map of each method using Grad-CAM~\cite{selvaraju2017grad}.
    These maps were visualized for the All-CNN model on the lacuna-10 dataset.
    }
    \vspace{-5mm}
    \label{fig:gradcam}
\end{figure}

\vspace{-4mm}
\paragraph{Visualization.}
To gain a clearer understanding of KD, we visualized the activation maps using Grad-CAM~\cite{selvaraju2017grad} on the facial domain dataset, Lacuna-10~\cite{golatkar2020eternal,kurmanji2023towards}. 
We trained an All-CNN model to classify identity and compared it with LAU, which is the most comparable method that also uses only the forgetting data. 
As shown in Figure~\ref{fig:gradcam}, the original model strongly focused on the target object in the image, particularly emphasizing facial areas to capture identity-related features. 
Notably, with \methodname{} and \refinemethod{}, we observed a distinct shift in activation from the subjects to the background, significantly more so compared to other baseline methods.
We also presented the results for other classical data, like car, along with additional t-SNE results, in the supplementary~\ref{supp:add_vis}.

\vspace{-2mm}
\section{Conclusion}
\vspace{-1mm}

In this paper, we advanced existing MU techniques based on real-world knowledge removal requests, thereby redefining the interpretation of MU. 
More importantly, we identified a critical issue of knowledge retention in features, and ultimately proposed \textit{Knowledge Deletion}.
In response to this, we introduced the \textit{Knowledge Retention Score}, an evaluation metric that measures the intensity of retained knowledge at the feature level after \taskname{}. 
This metric highlighted the limitations of existing methods and laid the foundation for advancing the \taskname{} field. 
To tackle this, we proposed \methodname{} and \refinemethod{}, which effectively remove the knowledge specified for forgetting in \taskname{}. 
Our methods, despite their simplicity, demonstrate significant performance improvements over existing works, achieving superior results in \taskname{} with minimal unlearning time. 
We hope our work inspires new insights and advances in the MU research community.

\section{Acknowledgement}
This work was supported by Korea Planning \& Evaluation Institute of Industrial Technology (KEIT) grant funded by the Korea government (MOTIE) (RS-2024-00444344), and in part by Institute of Information \& communications Technology Planning \& Evaluation (IITP) grant funded by the Korea government (MSIT) (No. RS-2019-II190079, Artificial Intelligence Graduate School Program (Korea University)).
{
    \small
    \bibliographystyle{ieeenat_fullname}
    \bibliography{main}
}

\appendix
\maketitlesupplementary

\section{Pesudo Code of \refinemethod{}}
\label{supp:algorithm}

\begin{algorithm} [!hb]
\caption{Erasing Space Concept with Training (ESC-T)}
\label{alg:mask}
\textbf{Input:} original feature extractor $h_{\psi}$, original classification head $g_{\phi}$, forgetting dataset $\mathcal{D}_f$, ground truth label $y$, and left singular vectors $\mathbf{U}$.\\
\textbf{Parameter:} number of epochs $T$, learning rate $\eta$, and threshold $\tau$.\\
\textbf{Output:}  learned binary mask $\mathbf{M}_R$.
\begin{algorithmic}[1]
\State Initialize $\mathbf{M}_0 = \mathbf{1} \leftarrow$ matrix of ones with the same size as left singular vectors $\mathbf{U}$.
\For{$t$ in $1...T$}
    \For{$(x, y) \in D_f$}
        \State $\mathbf{U}_{t-1}' = \mathbf{U} \odot \mathbf{M}_{t-1}$ 
        \Comment{Equation 5.}~~~
        \State $f_{t-1}(x)=g_\phi \circ h_{\psi_{t-1}}^{}(x) $ \Comment{Equation 6.}~~~
        \State $\mathcal{L}_{PCE}(f_{t-1}(x), y)$
        \Comment{Equation 4.}~~~
        \State $\mathbf{M}_{t} \leftarrow \mathbf{M}_{t-1} - \eta\nabla\mathcal{L}_{PCE}$
        \State $\mathbf{M}_{t} = \min\left(1, \max\left(0, \mathbf{M}_{t}\right)\right)$ 
        \newline \hspace*{2em} \Comment{Limit the mask to a range [0,1].}
    \EndFor
\EndFor
\State ${\mathbf{M}_{R}(i, j)}=\begin{cases}
    1, & \text{if ${\mathbf{M}_{R}(i, j)}>\tau$}.\\
    0, & \text{otherwise}.
 \end{cases}$
\end{algorithmic}
\end{algorithm}
\section{Additional Details}
\label{supp:add_details}

\subsection{Baselines}

\begin{itemize}
    \item \textbf{Negative Gradient (NG)}~\cite{thudi2022unrolling}: From the original model, NG performs gradient ascent on the forgetting data, which is in the opposite direction of the original model's training.
    \item \textbf{Random Label (RL)}~\cite{golatkar2020eternal}: RL starts from the original model and fine-tunes it in a manner similar to the original training process, \ie, using cross-entropy loss, but with randomly assigned labels. In our experiments, RL only used the randomly labeled forgetting data and did not utilize any remaining data. However, in the case of random data forgetting, RL also leverages the remaining data to maintain model performance.
    \item \textbf{Boundary Shrink (BS)}~\cite{chen2023boundary}: To advance RL, BS identifies the closest incorrect label for each forgetting sample based on an adversarial attack method and uses this label for unlearning, similar to the approach used in RL.
    \item \textbf{Layer Attack Unlearning (LAU)}~\cite{kim2023layer}: LAU uses Partial-PGD to perform adversarial attacks on the features of the forgetting data, training them to be predicted differently from the original model, while employing knowledge distillation to maintain decision boundaries for the remaining data.
    \item \textbf{Fisher}~\cite{golatkar2020eternal}: The Fisher (Forgetting) identifies the parameters that significantly influence the forgetting data and introduces noise to neutralize their effects from the original model.
    \item \textbf{BadT}~\cite{chundawat2023can}: BadT sets the original model as a competent teacher and the randomly initialized model as an incompetent teacher. The unlearned model is obtained by minimizing the KL divergence between its output and the competent teacher's output on the remaining data, while aligning with the incompetent teacher on the forgetting data.
    \item \textbf{SCRUB}~\cite{kurmanji2023towards}: SCRUB is also an unlearning method based on the distillation approach, which also utilizes the original model as a teacher. The objective of SCRUB is to minimize the KL divergence with the teacher on the remaining data while maximizing the KL divergence on the forgetting data. Additionally, cross-entropy loss is used on the remaining data to maintain model performance. 
    \item \textbf{$\ell_1$-sparse}~\cite{liu2023model}: $\ell_1$-sparse infuses weight sparsity into the unlearning process. The objective of $\ell_1$-sparse is to minimize the cross-entropy loss on the remaining data with an $\ell_1$ norm-based sparsity penalty. It directly removes knowledge from the feature extractor using a regularization term, achieving slightly better results in \taskname{} compared to existing methods. However, it necessarily requires remaining data for unlearning.
    \item \textbf{SalUn}~\cite{fan2024salun}: SalUn consists of a two-step process: finding the weight saliency map and performing the unlearning process. SalUn leverages the gradient-based weight saliency map to identify important parameters for unlearning using the NG loss. Based on this, SalUn updates only the top-$k$\% of parameters using any other forgetting loss, typically using the RL loss in practice.
\end{itemize}

\subsection{Linear Probing}

\begin{figure*}[!ht]
    \centering
    \begin{subfigure}[t]{.24\textwidth}
        \centering\includegraphics[width=\textwidth]{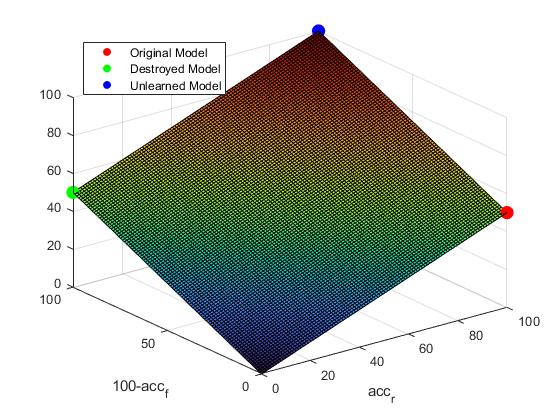}
        \caption{Arithmetic Mean\label{fig:metrics_a}}
    \end{subfigure}
    \begin{subfigure}[t]{.24\textwidth}
        \centering\includegraphics[width=\textwidth]{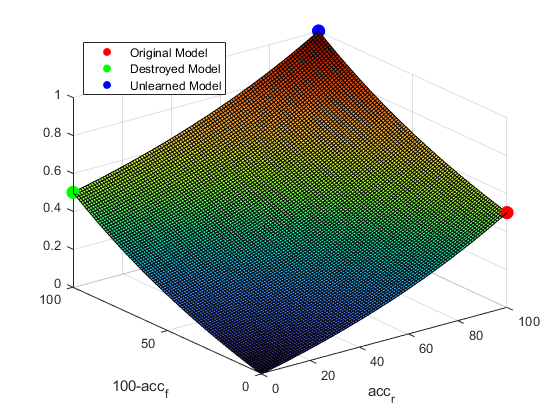}
        \caption{Unlearning Score~\cite{kim2023layer}}
    \end{subfigure}
    \begin{subfigure}[t]{.24\textwidth}
        \centering\includegraphics[width=\textwidth]{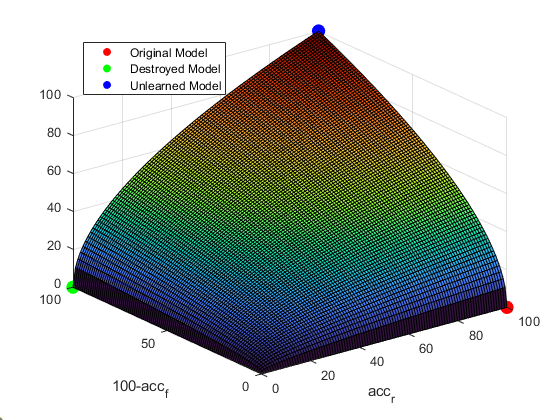}
        \caption{Geometric Mean}
    \end{subfigure}
    \begin{subfigure}[t]{.24\textwidth}
        \centering\includegraphics[width=\textwidth]{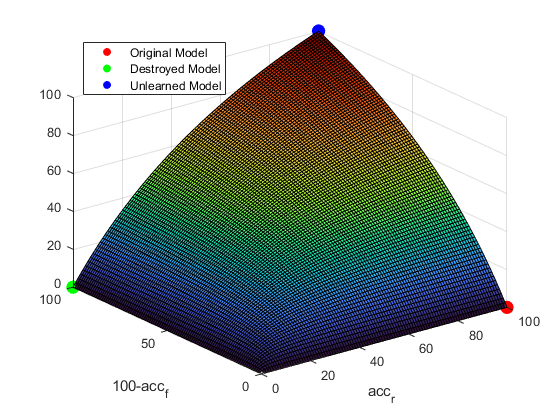}
        \caption{Harmonic Mean}
    \end{subfigure}

    \caption{Visualization of the distribution of various mean methods. The \textcolor{red}{red} dots represent the ideal original model, the \textcolor{green}{green} dots represent the destroyed model, and the \textcolor{blue}{blue} dots represent the completely unlearned model.}
    \label{fig:metrics}
\end{figure*}
\begin{table*}[!t]
    \centering
    \resizebox{0.9\textwidth}{!}{
    \begin{tabular}{cccc}
    \toprule
    Arithmetic Mean & Unlearning Score & Geometric Mean & Harmonic Mean  \\
    \midrule
    $\frac{acc_r+(100-acc_f)}{2}$ & $\frac{\text{exp}(\frac{acc_r}{100})+\text{exp}(1-\frac{acc_f}{100})-2}{2\cdot(\text{exp}(1)-1}$ & $\sqrt{acc_r\cdot(100-acc_f)}$ & $\frac{2 \cdot (100 - \text{acc}_f) \cdot \text{acc}_r}{(100 - \text{acc}_f) + \text{acc}_r}$ \\
    \bottomrule
    \end{tabular}
}
\caption{The equation of each types of mean methods.}
\label{tab:means_eq}
\end{table*}
\begin{table*}[!h]
    \centering
    \resizebox{0.7\textwidth}{!}{
    \begin{tabular}{lcccccc}
    \toprule
    \multirow{2}{*}{Method} & \multicolumn{6}{c}{CIFAR-100} \\ \cmidrule{2-7}
    & $D_{ft}(\downarrow)$ & $D_{rt}(\uparrow)$ & $AM_{\text{t}}(\uparrow)$ & $US_{\text{t}}(\uparrow)$ & $GM_{\text{t}}(\uparrow)$ & $HM_{\text{t}}(\uparrow)$ \\
    \midrule
    Original & 93.30 & 93.12 & 49.91 & 0.4676  & 24.98 & 12.50 \\
    Finetune~\cite{warnecke2021machine} & 42.20 & 90.89 & 74.35 & 0.6588  & 72.48 & 70.66 \\
    NG~\cite{thudi2022unrolling} & 87.80 & 92.44 & 52.32 & 0.4802  & 33.58 & 21.56 \\
    RL~\cite{golatkar2020eternal} & 13.60 & 23.46 & 54.93 & 0.4764  & 45.02 & 36.90 \\
    Boundary Shrink~\cite{chen2023boundary} & 7.60 & 24.17 & 58.29 & 0.5217  & 47.26 & 38.32 \\
    LAU~\cite{kim2023layer} & 0.00 & 93.13 & 96.57 & 0.9475  & 96.50 & 96.44 \\
    \textbf{\methodname{} (Ours)} & 0.10 & 93.22 & 96.56 & 0.9474 & 96.50 & 96.45\\
    \textbf{\refinemethod{} (Ours)} & 0.00 & 93.18 & 96.59 & 0.9479  & 96.53 & 96.47 \\
    \midrule
    Retrain & 0.00 & 93.16 & 96.58 & 0.9477  & 96.52 & 96.46 \\
    \bottomrule
    \end{tabular}
}
\caption{Comparison of various methods: Arithmetic Mean (AM), Unlearning Score (US), Geometric Mean (GM), and Harmonic Mean (HM).}
\label{tab:metrics}
\end{table*}

In Section 3.2.3., we introduced a novel benchmark~\metricname{} as an effective metric for evaluating the degree of~\taskname{}. 
\metricname{} performs linear probing using the feature extractor after \taskname{}, and then measures the utility, \ie, accuracy.
We conduct linear probing with frozen feature extractor of the unlearned model and randomly initialized classification head.
As is typically done, we only optimize the classification head with all training data \(D=D_f\cup D_r\).
We trained the classifier for 10 epochs using the SGD optimizer with a learning rate of 0.001 and a batch size of 64.
We also used linear probing in the experiments presented in Figure 1 of our main paper.

\section{Additional Discussion}

\subsection{\texorpdfstring{$HM$}{HM} Metric}

For a comprehensive evaluation, we employ the Harmonic Mean in our main paper.
An appropriate comprehensive evaluation can detect that the unlearned model can effectively remove the forget knowledge while preserving the remain knowledge.
To achieve this goal, we consider some failure cases of unlearning.
The original model should receive a low score in comprehensive evaluation because this model is completely not unlearned, and the destroyed model, which has accuracy 0\% in both remain (\(acc_r\)) and forget (\(acc_f\)) data in an extreme case, should also receive a low score.
On the other hand, ideal unlearned models, such as retrain model, should have high score in comprehensive evaluation.
Based on this case analysis, we seek the most appropriate mean method.
To find the most suitable method, we compare the Arithmetic Mean, Unlearning Score~\cite{kim2023layer}, Geometric Mean, and Harmonic Mean.
Each type of mean is calculated as illustrated in Table~\ref{tab:means_eq} and visualized in Figure~\ref{fig:metrics}.

\begin{table*}[!t]
    \centering
    \resizebox{\textwidth}{!}{
    \begin{tabular}{cccccgg}
    \toprule
    \multirow{2}{*}{\(p\)(\%)} & \multicolumn{6}{c}{CIFAR-100} \\ \cmidrule{2-7}
    &\(D_f(\downarrow)\)&\(D_r(\uparrow)\)&\(D_{ft}(\downarrow)\)&\(D_{rt}(\uparrow)\)&\(HM(\uparrow)\)&\(HM_{\text{t}}(\uparrow)\)\\
    \midrule
        1  & 35.20  & 98.13  & 32.50  & 93.09  & 78.06  & 78.26 \\
        \textbf{1.7} & 0.02 & 98.05 & 0.10 & 93.22 & \textbf{99.01} & \textbf{96.45} \\
        5 & 0.22 & 97.41 & 0.10 & 92.40 & 98.58 & 96.00 \\
        {10}  & 0.16  & 96.81  & 0.10  & 91.53  & {98.30}  & {95.53} \\
        30  & 0.12  & 94.23  & 0.10  & 89.38  & 96.97  & 94.35 \\
        50  & 0.16  & 91.99  & 0.10  & 86.97  & 95.75  & 92.99 \\
        70  & 0.02  & 84.11  & 0.00  & 79.63  & 91.36  & 88.66 \\
        90  & 0.02  & 44.77  & 0.00  & 42.38  & 61.85  & 59.53 \\
    \bottomrule
    \end{tabular}
    \begin{tabular}{ccccgg}
    \toprule
    \multicolumn{6}{c}{CIFAR-100$-KR$} \\ \cmidrule{1-6}
   \(D_f(\downarrow)\)&\(D_r(\uparrow)\)&\(D_{ft}(\downarrow)\)&\(D_{rt}(\uparrow)\)&\(HM(\uparrow)\)&\(HM_{\text{t}}(\uparrow)\)\\
    \midrule
        34.42  & 97.00  & 33.30  & 92.12  & 78.25  & 77.38 \\ 
        0.36 & 97.00 & 0.40 & 92.09 & \textbf{98.30} & \textbf{95.70} \\ 
        0.42 & 95.13 & 0.40 & 90.24 & 97.30 & 94.69 \\ 
        0.44 & 91.18 & 0.30 & 86.47 & {95.19}  & {92.61} \\ 
        0.90 & 77.34 & 0.90 & 73.71 & 86.88  & 84.54 \\ 
        0.80 & 62.46 & 0.80 & 59.07 & 76.66  & 74.05 \\ 
        1.02 & 37.23 & 1.10 & 35.30 & 54.11  & 52.03 \\ 
        0.78 & 7.89  & 1.30 & 7.30  & 14.62  & 13.59 \\ 
    \bottomrule
    \end{tabular}
    \vspace{-2mm}   
}
\caption{Ablation study of hyperparameter \(p\) in \methodname{}. We report accuracy and KR using ViT. The results show that our methods are robust to various values of the hyperparameter \(p\). However, if \(p\) is too large, \methodname{} results in over-deletion, causing both the forgetting knowledge and some of the remaining knowledge to be lost.
}
\label{tab:full_p}
\end{table*}

In the first case, the Arithmetic Mean increases with larger \(acc_{r}\) and smaller \(acc_{f}\) (Figure~\ref{fig:metrics} (a)).
However, it remains fixed at 50, even when the model exhibits complete failure with 0\% \(acc_{r}\) and 0\% \(acc_{f}\), or in scenarios where unlearning completely fails, resulting in \(acc_{r}\) and \(acc_{f}\) both being 100\%.
These characteristics mean that it is not a good metric for evaluating unlearning as we mentioned.
Unlearning Score~\cite{kim2023layer} also has the same problem because it applies an exponential to \(acc\) but follows the Arithmetic Mean (Figure~\ref{fig:metrics} (b)).
To complement this, we can think of Geometric Mean and Harmonic Mean (Figure~\ref{fig:metrics} (c), (d)).
Both have a value of zero in situations where the model is completely broken or not unlearned, making them ideal for evaluating unlearning.
However, as the Harmonic Mean is more sensitive to small values compared to the Geometric Mean, it yields a lower value (12.5) than the Geometric Mean (24.98) when the model is the original model.
Therefore, it is better suited for comprehensive evaluation in unlearning scenarios according to all casess.
Table~\ref{tab:metrics} shows that when applying the metrics to an unlearning experiment with Vision Transformer~(ViT)~\cite{dosovitskiy2021image} on CIFAR-100~\cite{krizhevsky2009learning}, the Harmonic Mean is the best at differentiating small values.
The Harmonic Mean reflects the characteristics of unlearning well, with the original model being close to 0\% and the unlearned model being close to 100\%.

\subsection{What is Knowledge in Deep Learning Model?}
Our methods are designed to remove forgetting knowledge from the original model, even at the feature level. 
In this section, we further discuss the nature of knowledge in deep learning models. 
Deep learning models accumulate learned knowledge as data passes through the network, with the embedding features containing the most accumulated knowledge.
This accumulated knowledge consists of sub-features that represent lower-level knowledge compared to the target knowledge, \ie, class.

For example, the identity of person A might be defined by a combination of an oval face shape, straight eyebrows, blue eyes, and a Roman nose. 
Similarly, the identity of person B could be characterized by a long face shape, arched eyebrows, black eyes, and a Nubian nose. 
While these sub-features may be present across different classes and can be captured from the remaining knowledge, it is the specific combination of these features that allows the model to identify particular knowledge.
As mentioned in Section 2.3.1., even if the model operates based on the the remaining knowledge, this is not sufficient for \taskname{}.
Therefore, it is crucial to remove the model's ability to effectively combine these features for identification.
This is why our methods also outperform the retrained model in KR.


\subsection{Setting of Random Data Forgetting}

We present the random data forgetting experiments in Table 5 of our main paper.
The goal of random data forgetting is to remove the influence of specific data from the original model, and the optimal point for achieving this is the retrained model.
This is distinct from our primary focus in \taskname{}, which involves knowledge removal requests from users.
For this reason, unlike \taskname{}, we evaluate the effectiveness of random data forgetting by comparing it to the retrain-from-scratch model.
Furthermore, we used the average gap as an integrated value for each metric, similar to~\cite{liu2023model, fan2024salun}.

\section{Additional Ablation}
\label{supp:add_abl}

\subsection{\texorpdfstring{$p$}{p} Ablation}

Table~\ref{tab:full_p} shows the full results of the hyperparameter \(p\) ablation study.
This shows that as \(p\) gets larger, more knowledge is erased.
When we erased very little, like 1\%, model retained a lot of forget knowledge, and when we erased a lot, like 90\%, model lost a lot of remain knowledge.
\methodname{} achieves relatively robust results across various values of \(p\).
Despite the performance is good for most $p$, we empirically found that \methodname{} achieves the most stable results around 1.7\%.
For this reason, we selected the pruning hyperparameter $p$ between 1 and 3 in our experiments. 
Some variation is needed because each model has a different embedding dimension size, and each principal direction contains varying levels of knowledge intensity.
However, we consistently obtain comparable results for any value of $p$ within this range.

\subsection{The Effect of Clipping}
\label{sup:clipping}

\begin{figure*}[!t]
    \centering
    \begin{subfigure}[t]{.46\textwidth}
        \centering\includegraphics[width=\textwidth]{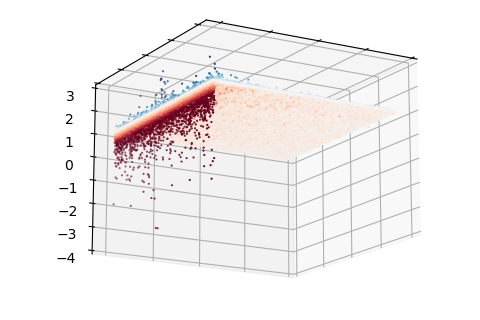}
        \caption{Updated mask without clipping.}
    \end{subfigure}
    \begin{subfigure}[t]{.46\textwidth}
        \centering\includegraphics[width=\textwidth]{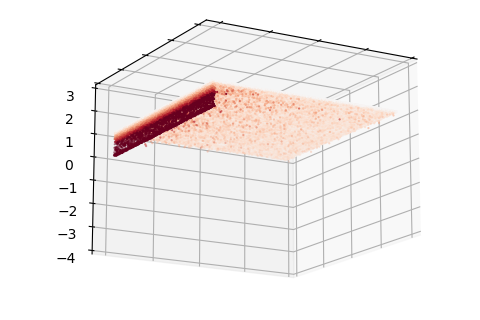}
        \caption{Updated mask with clipping.}
    \end{subfigure}
    \begin{subfigure}[t]{.46\textwidth}
        \centering\includegraphics[width=\textwidth]{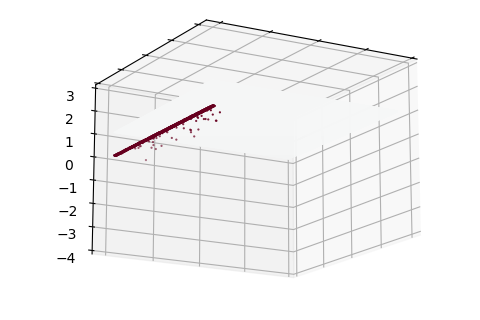}
        \caption{After threshold without clipping.}
    \end{subfigure}
    \begin{subfigure}[t]{.46\textwidth}
        \centering\includegraphics[width=\textwidth]{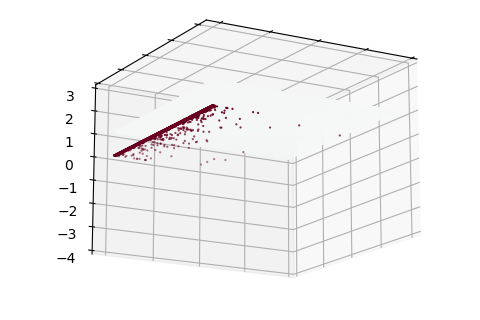}
        \caption{After threshold with clipping.}
    \end{subfigure}
    \caption{Visualization of clipping effect on the unlearned mask after training and the threshold. The results show that the unlearned mask utilizes more elements when the clipping method is employed. This also means that more elements can be used, after threshold, for unlearning, allowing \refinemethod{} to efficiently remove more forget knowledge while preserving the remaining knowledge.}
    \label{fig:mask}
\end{figure*}
In our additional experiments from Table 3 in our main paper, we investigated the impact of restricting masks to values between 0 and 1 during the update process in \refinemethod{}.
As we mentioned in Section 4.3. in our main paper, when we introduced clipping to confine mask values within the range of 0 to 1, we observed an improvement in KR performance.
To understand this, we visualized the unlearned mask \(\textbf{M}_T\) in Figure~\ref{fig:mask}.
As shown in Figure~\ref{fig:mask} (a), without clipping, some elements are updated to have out of range values, leading to overconfidence about forgetting data in the classification head. 
This overconfidence triggers an early stopping of our training process, as it meets the criterion of our strategy where all forgetting data are misclassified.
Consequently, as shown in Figure~\ref{fig:mask} (c), this approach fails to fully evaluate the influence of each element, resulting in the missing of crucial elements pertinent to the \taskname{} of forgetting data.
On the other hand, with clipping, our training process relatively considers the full range of mask as shown in Figure~\ref{fig:mask} (b) and (d).
Clipping makes it possible for \refinemethod{} to capture more important elements, enabling us to effectively eliminate forgetting knowledge.
Furthermore, these results demonstrate the validity of our design choice for \methodname{}, as only a few directional elements play crucial roles for unlearning process.

We presented the mask after threshold in Figure~\ref{fig:mask} (c) and (d).
Threshold ensures that crucial forgetting knowledge is completely erased (0), while less important knowledge is fully retained (1), allowing for the preservation of remaining knowledge.
This clear distinction based on importance significantly enhances the performance of KR, leading to improved outcomes.
By applying clipping and threshold together, we can get the optimal KR performance while maintaining good utility.
\section{Additional Experiments}
\label{supp:add_exp}

\subsection{Additional Results for KD}

\begin{table*}[tb]
\centering
\resizebox{\textwidth}{!}{
\begin{tabular}{lccccgg}
\toprule
\multirow{2}{*}{Method} & \multicolumn{6}{c}{CIFAR-100} \\ \cmidrule{2-7}
& $D_f(\downarrow)$ & $D_r(\uparrow)$ & $D_{ft}(\downarrow)$ & $D_{rt}(\uparrow)$ & $HM(\uparrow)$ & $HM_{\text{t}}(\uparrow)$ \\
\midrule
Original & 98.42 & 98.10 & 93.30 & 93.12 & 3.11 & 12.50\\
Finetune~\cite{warnecke2021machine} & 43.60 & 98.31 & 42.20 & 90.89 & 71.68 & 70.66\\
NG~\cite{thudi2022unrolling} & 46.92 & 66.87 & 42.60 & 64.30 & 59.18 & 60.65 \\
RL~\cite{golatkar2020eternal} & 14.58 & 24.73 & 13.60 & 23.46 & 38.36 & 36.90\\
Boundary Shrink~\cite{chen2023boundary} & 8.26 & 25.06 & 7.60 & 24.17 & 39.37 & 38.32\\
LAU~\cite{kim2023layer} & 0.00 & 97.99 & 0.00 & 93.13 & 98.98 & 96.44 \\
\textbf{\methodname{} (Ours)} & 0.02 & 98.05 & 0.10 & 93.22 & \underline{99.01} & \underline{96.45} \\
\textbf{\refinemethod{} (Ours)} & 0.00 & 98.11 & 0.00 & 93.18 & \textbf{99.05} & \textbf{96.47}\\
\midrule
Retrain & 0.00 & 98.48 & 0.00 & 93.16 & 99.23 & 96.46\\
\bottomrule
\end{tabular}

\begin{tabular}{ccccgg}
\toprule
\multicolumn{6}{c}{CIFAR-100$-KR$} \\ \cmidrule{1-6}
$D_f(\downarrow)$ & $D_r(\uparrow)$ & $D_{ft}(\downarrow)$ & $D_{rt}(\uparrow)$ & $HM(\uparrow)$ & $HM_{\text{t}}(\uparrow)$ \\
\midrule
97.78 & 97.01 & 91.70 & 91.83 & 4.34 & 15.22\\
76.14 & 96.62 & 70.70 & 89.11 & 38.27 & 44.10\\
95.14 & 96.32 & 88.20 & 91.41 & 9.25 & 20.90\\
11.76 & 17.04 & 10.80 & 16.47 & 28.56 & 27.81\\
19.30 & 30.21 & 17.10 & 29.96 & 43.96 & 44.01\\
97.78 & 97.01 & 91.70 & 91.83 & 4.34 & 15.22\\
0.36 & 97.00 & 0.40 & 92.09 & \textbf{98.30} & \textbf{95.70} \\
1.80 & 96.93 & 1.60 & 92.04 & \underline{97.56} & \underline{95.11}\\
\midrule
71.34 & 97.17 & 71.20 & 91.90 & 44.26 & 43.86\\
\bottomrule
\end{tabular}
}
\caption{Accuracy and KR using ViT on CIFAR-100.
Bold at the best value and underlined at the second.}
\label{tab:vit_cifar100}
\end{table*}

\begin{table*}[!t]
\vspace{-3mm}
\centering
\resizebox{\textwidth}{!}{
\begin{tabular}{lccccgg}
\toprule
\multirow{2}{*}{Method} & \multicolumn{6}{c}{Lacuna-10} \\ \cmidrule{2-7}
& $D_f(\downarrow)$ & $D_r(\uparrow)$ & $D_{ft}(\downarrow)$ & $D_{rt}(\uparrow)$ & $HM(\uparrow)$ & $HM_{\text{t}}(\uparrow)$ \\
\midrule
Original & 100.0 & 100.0 & 92.00 & 86.56 & 0.00 & 14.65 \\
LAU~\cite{kim2023layer} & 0.00 & 100.0 & 0.00 & 86.22 & 100.0 & 92.60 \\
\textbf{\methodname{} (Ours)} & 1.50 & 100.0 & 0.00 & 88.00 & 99.24 & 93.62 \\
\textbf{\refinemethod{} (Ours)} & 0.00 & 100.0 & 0.00 & 88.11 & \textbf{100.0} & \textbf{93.68} \\
\midrule
Retrain & 0.00 & 100.0 & 0.00 & 89.33 & 100.0 & 94.36 \\
\bottomrule
\end{tabular}

\begin{tabular}{ccccgg}
\toprule
\multicolumn{6}{c}{Lacuna-10$-KR$} \\ \cmidrule{1-6}
$D_f(\downarrow)$ & $D_r(\uparrow)$ & $D_{ft}(\downarrow)$ & $D_{rt}(\uparrow)$ & $HM(\uparrow)$ & $HM_{\text{t}}(\uparrow)$ \\
\midrule
100.0 & 100.0 & 89.00 & 85.89 & 0.00 & 19.50 \\
100.0 & 100.0 & 89.00 & 85.89 & 0.00 & 19.50 \\
12.25 & 100.0 & 10.00 & 87.22 & \textbf{93.48} & 88.59 \\
12.25 & 100.0 & 9.00 & 87.22 & \textbf{93.48} & \textbf{89.07} \\
 \midrule
81.00 & 100.0 & 64.00 & 84.78 & 91.93 & 50.54 \\   
\bottomrule
\end{tabular}
}
\caption{Accuracy and KR using All-CNN on Lacuna-10.}
\label{tab:lacuna_allcnn}
\end{table*}

\begin{table*}[!t]
\vspace{-3mm}
\centering
\resizebox{\textwidth}{!}{
\begin{tabular}{lccccgg}
\toprule
\multirow{2}{*}{Method} & \multicolumn{6}{c}{Lacuna-100} \\ \cmidrule{2-7}
& $D_f(\downarrow)$ & $D_r(\uparrow)$ & $D_{ft}(\downarrow)$ & $D_{rt}(\uparrow)$ & $HM(\uparrow)$ & $HM_{\text{t}}(\uparrow)$ \\
\midrule
Original & 94.58 & 94.88 & 90.60 & 90.30 & 10.25 & 17.03 \\
LAU~\cite{kim2023layer} & 0.00 & 88.55 & 0.00 & 83.97 & 93.93 & 91.29\\
\textbf{\methodname{} (Ours)} & 0.40 & 86.54 & 0.10 & 81.09 & 92.61 & 89.52 \\
\textbf{\refinemethod{} (Ours)} & 0.00 & 93.38 & 0.00 & 89.87 & 96.58 & 94.66\\
\midrule
Retrain & 40.53 & 96.24 & 38.60 & 91.30 & 73.51 & 73.42 \\
\bottomrule
\end{tabular}

\begin{tabular}{ccccgg}
\toprule
\multicolumn{6}{c}{Lacuna-100$-KR$} \\ \cmidrule{1-6}
$D_f(\downarrow)$ & $D_r(\uparrow)$ & $D_{ft}(\downarrow)$ & $D_{rt}(\uparrow)$ & $HM(\uparrow)$ & $HM_{\text{t}}(\uparrow)$ \\
\midrule

95.48 & 94.98 & 93.10 & 90.61 & 8.63 & 12.82\\
95.48 & 94.98 & 93.10 & 90.61 & 8.63 & 12.82\\
0.10 & 89.65 & 0.40 & 85.10 & 94.50 & 91.78 \\
58.68 & 94.69 & 54.90 & 90.50 & 57.53 & 60.20 \\
 \midrule
92.05 & 95.42 & 89.20 & 91.27 & 13.91 & 19.31 \\   
\bottomrule
\end{tabular}
}
\caption{Accuracy and KR using ViT on Lacuna-100.}
\label{tab:lacuna_vit}
\end{table*}

\paragraph{Additional in ViT.}
We also conducted the \taskname{} with ViT on the CIFAR-100 dataset, and Table~\ref{tab:vit_cifar100} provides the results.
In this experiments, we utilized the ViT and removed 10 classes from the trained model.
Except the LAU, they suffer from the over-forgetting or under-forgetting in the utility, and feature-level knowledge deletion is also restricted.
LAU has comparable results in utility, but it also fail to remove the feature knowledge. 
In contrast, our methods were still outperform to other comparisons both utility and KR.

\paragraph{Experiments of Facial Domains.}
To further validate and strengthen the concept of \taskname{}, we conducted experiments with facial domain datasets.
Given the risks associated with extensive personal data in face recognition systems, our experiments with these datasets are particularly relevant. 
Lacuna-10~\cite{golatkar2020eternal,kurmanji2023towards} consists of face images of 10 celebrities from VGGFaces2~\cite{cao2018vggface2}, randomly sampled with at least 500 images each. 
It was split into a test set of 100 samples per class, while the remaining samples formed the training set. 
In Figure 6 of our main paper, we visualize the effectiveness of our methods using Grad-CAM on this dataset.
In Table~\ref{tab:lacuna_allcnn}, we present the quantitative results of these experiments. 
Our methods continue to demonstrate strong effectiveness for \taskname{} in the facial domain, successfully mitigating privacy concerns while maintaining model performance.
In addition, we also conducted experiments with a larger facial domain dataset, Lacuna-100, which contains randomly sampled images of 100 celebrities with at least 500 images each. 
We selected the ViT model as it is more appropriate for the larger dataset, and the results are shown in Table~\ref{tab:lacuna_vit}.
These results again highlight the effectiveness of our proposed methods in protecting individual privacy by removing specific identity knowledge from the trained models.

\begin{table*}[!t]
\label{tab:full_imagenet}
\centering
\resizebox{\textwidth}{!}{
\begin{tabular}{lccccgg}
\toprule
\multirow{2}{*}{Method} & \multicolumn{6}{c}{ImageNet-1K} \\ \cmidrule{2-7}
& $D_f(\downarrow)$ & $D_r(\uparrow)$ & $D_{ft}(\downarrow)$ & $D_{rt}(\uparrow)$ & $HM(\uparrow)$ & $HM_{\text{t}}(\uparrow)$ \\
\midrule
Original & 87.80 & 87.40 & 80.90 & 80.25 & 21.41 & 30.86 \\
LAU~\cite{kim2023layer} & 0.00 & 67.01 & 0.00 & 62.09 & 80.25 & 76.61 \\
\textbf{\methodname{} (Ours)} & 0.11 & 83.81 & 0.06 & 79.15 & \underline{91.15} & \underline{88.34} \\
\textbf{\refinemethod{} (Ours)} & 0.01 & 84.81 & 0.00 & 80.14 & \textbf{91.78} & \textbf{88.98} \\
\bottomrule
\end{tabular}

\begin{tabular}{ccccgg}
\toprule
\multicolumn{6}{c}{ImageNet-1K$-KR$} \\ \cmidrule{1-6}
$D_f(\downarrow)$ & $D_r(\uparrow)$ & $D_{ft}(\downarrow)$ & $D_{rt}(\uparrow)$ & $HM(\uparrow)$ & $HM_{\text{t}}(\uparrow)$ \\
\midrule
87.18 & 86.92 & 80.96 & 79.78 & 22.34 & 30.74 \\
87.18 & 86.92 & 80.96 & 79.78 & 22.34 & 30.74 \\
0.04 & 84.72 & 0.00 & 80.15 & \textbf{91.71} & \textbf{88.98} \\
56.97 & 86.35 & 50.40 & 80.27 & \underline{57.44} & \underline{61.31} \\
\bottomrule
\end{tabular}
}
\caption{Accuracy and KR using ViT on ImageNet-1k.
Bold at the best and underlined at the second.}
\label{tab:imagenet1k}
\end{table*}

\paragraph{Large Scale Dataset.}
Our methods are simple and significantly efficient for \taskname{}. 
For this reason, they can be easily applied to larger-scale datasets, such as ImageNet-1K~\cite{russakovsky2015imagenet}. 
In Table~\ref{tab:imagenet1k}, our methods achieve superior performance compared to LAU in both utility and KR. 
These results indicate that our methods are robust to dataset size, making them highly useful in real-world applications, where most deep learning models are built on large-scale datasets.
In these experiments, we only use LAU for comparison because it is the most efficient method among the existing ones.

\begin{table*}[!t]
\vspace{-3mm}
\centering
\resizebox{\textwidth}{!}{
\begin{tabular}{lccccgg}
\toprule
\multirow{2}{*}{Method} & \multicolumn{6}{c}{CUB-200-2011} \\ \cmidrule{2-7}
& $D_f(\downarrow)$ & $D_r(\uparrow)$ & $D_{ft}(\downarrow)$ & $D_{rt}(\uparrow)$ & $HM(\uparrow)$ & $HM_{\text{t}}(\uparrow)$ \\
\midrule
Original & 94.82 & 95.66 & 83.81 & 85.03 & 9.82 & 27.20 \\
LAU~\cite{kim2023layer} & 0.00 & 92.90 & 0.18 & 82.51 & 96.62 & 90.34 \\
\textbf{\methodname{} (Ours)} & 0.00 & 89.51 & 0.00 & 79.19 & 94.46 & 88.39 \\
\textbf{\refinemethod{} (Ours)} & 0.00 & 92.55 & 0.00 & 82.72 & 96.13 & 90.54\\
\midrule
Retrain & 48.25 & 98.74 & 45.86 & 86.18 & 67.91 & 66.50 \\
\bottomrule
\end{tabular}

\begin{tabular}{ccccgg}
\toprule
\multicolumn{6}{c}{CUB-200-2011$-KR$} \\ \cmidrule{1-6}
$D_f(\downarrow)$ & $D_r(\uparrow)$ & $D_{ft}(\downarrow)$ & $D_{rt}(\uparrow)$ & $HM(\uparrow)$ & $HM_{\text{t}}(\uparrow)$ \\
\midrule
86.81 & 92.60 & 78.42 & 82.04 & 23.09 & 34.17 \\
86.81 & 92.60 & 78.42 & 82.04 & 23.09 & 34.17 \\
0.83 & 76.18 & 0.18 & 68.52 & 86.17 & 81.26 \\
1.50 & 85.73 & 1.08 & 77.05 & 91.97 & 86.63 \\
 \midrule
76.79 & 93.46 & 69.24 & 80.53 & 37.19 & 44.42 \\   
\bottomrule
\end{tabular}
}
\caption{Accuracy and KR using ViT on CUB-200-2011.}
\label{tab:cub_vit}
\end{table*}

\begin{figure*}[!ht]
    \centering
    \includegraphics[width=\textwidth]{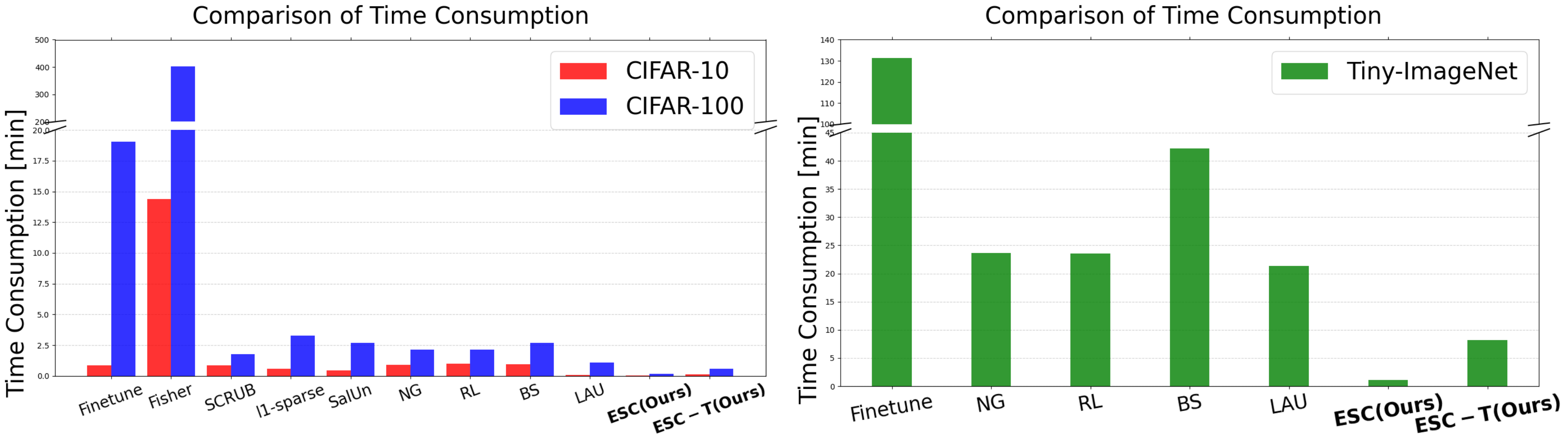}
    \caption{Comparison of time consumption. In addition to Figure 5 of our main paper, we present further results from other comparisons using the remaining data on the left. On the right side, we also illustrate the results on the Tiny-ImageNet dataset.}
    \label{fig:time_supp} 
\end{figure*}

\paragraph{Fine-Grained Dataset.}
In real-world scenario, deep learning model need to consider fine-grained datasets, distinguishing subtle differences between objects is often necessary.
Such datasets enable the model to learn complex objects with high precision, ensuring reliable performance that aligns with real-world needs. 
Furthermore, \taskname{} must effectively address these demands by appropriately handling subtle differences between objects.
This is challenging because the objects share much of the general knowledge, differing only in fine details. 
To verify our methods in this setting, we conducted \taskname{} using the CUB-200-2011 dataset~\cite{wah2011caltech}, which includes 200 different bird species, each annotated in detail. 
Approximately 60 images represent each species, creating a diverse and rich dataset. 
Despite these challenges, the results demonstrated the generalizability of our methods in a fine-grained setting.

\subsection{Additional Time Consumption}
To extend Figure 5 from our main paper, we present the time consumption for all comparisons on the CIFAR-10 and CIFAR-100 datasets in Figure~\ref{fig:time_supp}. 
When using the remaining data, significantly more time is typically required for \taskname{} since the remaining data is much larger than the forgetting data (almost nine times larger).
In contrast, \methodname{} requires only a single forward pass, achieving the lowest time consumption. 
\refinemethod{} also involves lightweight training, updating only the mask, making it faster than other methods.
This efficiency becomes even more remarkable as the dataset and model size grow.
As shown in Figure~\ref{fig:time_supp} left, where both the model (ViT) and dataset (Tiny-ImageNet) are larger, the efficiency of our methods is significantly higher than other methods, with a difference of more than 10 minutes.

\begin{figure*}[!t]
    \centering
    \includegraphics[width=\textwidth]{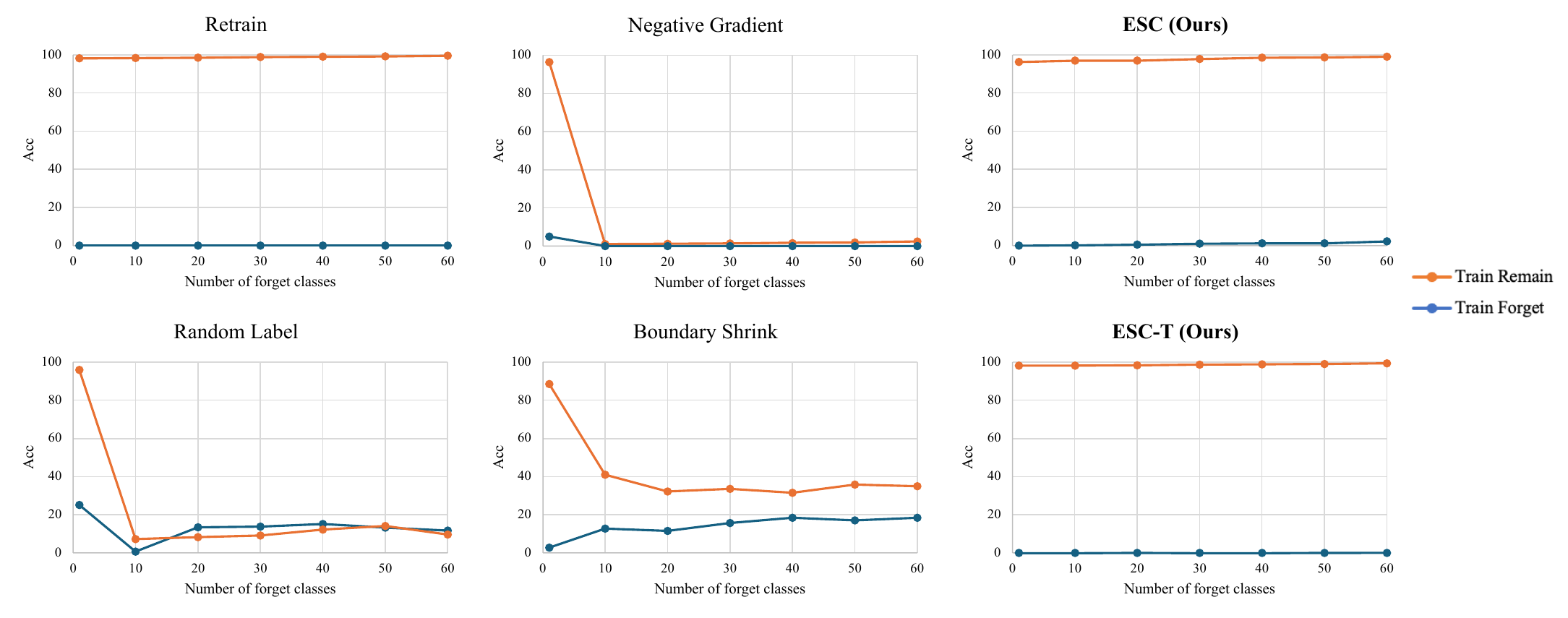}
    \caption{Multi-class unlearning experiments on diverse baselines. We use the results of train accuracy on CIFAR-100 with ViT.}
    \label{fig:multi-class}
    \vspace{-2mm}
\end{figure*}
\begin{figure*}[!t]
    \centering
    \includegraphics[width=0.9\textwidth]{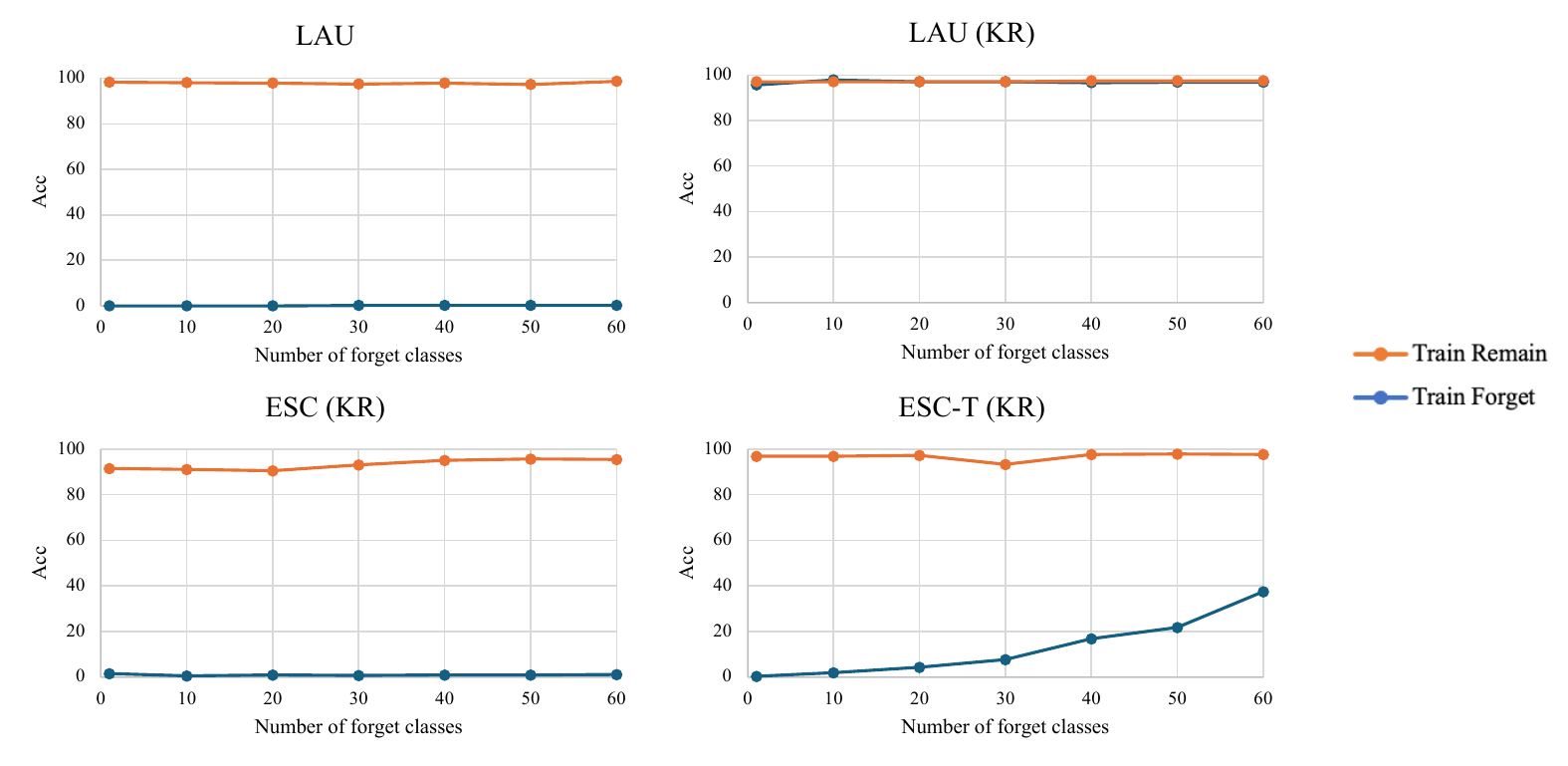}
    \caption{Multi-class unlearning experiments on LAU~\cite{kim2023layer} and ours. While LAU performs well in utility, it completely recover the forgetting knowledge in KR. In contrast, \methodname{} and \refinemethod{} still preserve the performance in KR.}
    \label{fig:multi-class2}
\end{figure*}
\subsection{Analysis by the Number of Target Classes}

In this section, we explore the impact of the number of forgetting classes on \taskname{} performance.
We conducted this experiment on the CIFAR-100 with ViT, varying the number of forgetting classes from 1 to 60.
In Figure~\ref{fig:multi-class}, we visualized the variation of accuracy with other baselines that exhibit poor multi-class unlearning performance, as discussed in our main paper.
For all baselines, we searched for the best hyperparameter for single-class unlearning and adopted this for other multi-class unlearning scenarios.
Our methods use the same hyperparameter settings as detailed in our main paper.

\begin{table*}[tb]
\centering
\resizebox{\textwidth}{!}{
\begin{tabular}{lccccgg}
\toprule
\multirow{2}{*}{Method} & \multicolumn{6}{c}{CIFAR-10} \\ \cmidrule{2-7}
& $D_f(\downarrow)$ & $D_r(\uparrow)$ & $D_{ft}(\downarrow)$ & $D_{rt}(\uparrow)$ & $HM(\uparrow)$ & $HM_{\text{t}}(\uparrow)$ \\
\midrule
Original & 99.96 & 99.91 & 99.20 & 99.02 & 0.08 & 1.59 \\
\methodname (Ours) & 1.92 & 99.56 & 2.30 & 98.36 & 98.81 & 98.03 \\
\refinemethod (Ours) & 0.86 & 99.92 & 0.50 & 99.06 & 99.53 & 99.28 \\
\bottomrule
\end{tabular}

\begin{tabular}{ccccgg}
\toprule
\multicolumn{6}{c}{CIFAR-10$-KR$} \\ \cmidrule{1-6}
$D_f(\downarrow)$ & $D_r(\uparrow)$ & $D_{ft}(\downarrow)$ & $D_{rt}(\uparrow)$ & $HM(\uparrow)$ & $HM_{\text{t}}(\uparrow)$ \\
\midrule
99.98 & 99.89 & 99.40 & 98.92 & 0.04 & 1.19 \\
73.48 & 99.72 & 70.40 & 98.63 & 41.90 & 45.53\\
82.84 & 99.90 & 81.70 & 98.96 & 29.29 & 30.89\\
\bottomrule
\end{tabular}
}

\resizebox{\textwidth}{!}{
\begin{tabular}{lccccgg}
\toprule
\multirow{2}{*}{Method} & \multicolumn{6}{c}{CIFAR-100} \\ \cmidrule{2-7}
& $D_f(\downarrow)$ & $D_r(\uparrow)$ & $D_{ft}(\downarrow)$ & $D_{rt}(\uparrow)$ & $HM(\uparrow)$ & $HM_{\text{t}}(\uparrow)$ \\
\midrule
Original & 98.42 & 98.10 & 93.30 & 93.12 & 3.11 & 12.50\\
\methodname (Ours) & 0.20 & 95.61 & 0.00 & 90.67 & 97.66 & 95.11 \\
\refinemethod (Ours) & 0.98 & 98.01 & 0.50 & 93.09 & 98.51 & 96.19 \\
\bottomrule
\end{tabular}
\begin{tabular}{ccccgg}
\toprule
\multicolumn{6}{c}{CIFAR-100$-KR$} \\ \cmidrule{1-6}
$D_f(\downarrow)$ & $D_r(\uparrow)$ & $D_{ft}(\downarrow)$ & $D_{rt}(\uparrow)$ & $HM(\uparrow)$ & $HM_{\text{t}}(\uparrow)$ \\
\midrule
97.78 & 97.01 & 91.70 & 91.83 & 4.34 & 15.22 \\
31.70 & 97.93 & 28.30 & 86.74 & 80.47 & 78.51\\
40.64 & 96.54 & 36.70 & 92.01 & 73.52 & 75.00 \\
\bottomrule
\end{tabular}
}

\resizebox{\textwidth}{!}{
\begin{tabular}{lccccgg}
\toprule
\multirow{2}{*}{Method} & \multicolumn{6}{c}{Tiny-ImageNet} \\ \cmidrule{2-7}
& $D_f(\downarrow)$ & $D_r(\uparrow)$ & $D_{ft}(\downarrow)$ & $D_{rt}(\uparrow)$ & $HM(\uparrow)$ & $HM_{\text{t}}(\uparrow)$ \\
\midrule
Original & 96.72 & 96.47 & 90.30 & 90.24 & 6.34 & 17.52\\
\methodname (Ours) & 0.12 & 94.94 & 0.00 & 89.36 & 97.35 & 94.38 \\
\refinemethod (Ours) & 0.14 & 96.40  & 0.00 & 90.51 & 98.10 & 95.02\\
\bottomrule
\end{tabular}

\begin{tabular}{ccccgg}
\toprule
\multicolumn{6}{c}{Tiny-ImageNet$-KR$} \\ \cmidrule{1-6}
$D_f(\downarrow)$ & $D_r(\uparrow)$ & $D_{ft}(\downarrow)$ & $D_{rt}(\uparrow)$ & $HM(\uparrow)$ & $HM_{\text{t}}(\uparrow)$ \\
\midrule
95.55 & 94.80 & 90.50 & 88.57 & 8.50 & 17.16 \\
1.01 & 94.76 & 0.40 & 89.12 & 96.83 & 94.07\\
7.36 & 94.58 & 6.30 & 88.78 & 93.60 & 91.17\\
\bottomrule
\end{tabular}
}
\caption{Accuracy and KR performance in ViT, using unseen data (test data) for KD. Our methods shows comparable results only with unseen data on various datasets.}
\label{tab:unseen}
\end{table*}

\begin{table*}[!t]
\centering
\resizebox{\textwidth}{!}{
\begin{tabular}{lcccccgg}
\toprule
\multirow{2}{*}{Method} & \multirow{2}{*}{\# of sequence} & \multicolumn{6}{c}{CIFAR-100} \\ \cmidrule{3-8}
& & $D_f(\downarrow)$ & $D_r(\uparrow)$ & $D_{ft}(\downarrow)$ & $D_{rt}(\uparrow)$ & $HM(\uparrow)$ & $HM_{\text{t}}(\uparrow)$ \\
\midrule
\multirow{4}{*}{\methodname{} (Ours)} & 1 & 0.02 & 98.05 & 0.10 & 93.22 & 99.01 & 96.45 \\
& 2 & 0.04 & 98.04 & 0.20 & 93.04 & 98.99 & 96.30 \\
& 5 & 0.16 & 97.34 & 0.50 & 92.30 & 98.57 & 95.76 \\
& 10 & 1.58 & 97.87 & 1.30 & 93.00 & 98.14 & 95.77 \\
\midrule
\multirow{4}{*}{\refinemethod{} (Ours)} & 1 & 0.00 & 98.11 & 0.00 & 93.18 & 99.05 & 96.47 \\
& 2 & 0.00 & 98.09 & 0.00 & 93.19 & 99.04 & 96.47 \\
& 5 & 0.00 & 98.09 & 0.00 & 93.14 & 99.04 & 96.45 \\
& 10 & 0.04 & 98.10 & 0.00 & 93.2 & 99.02 & 96.48 \\

\bottomrule
\end{tabular}
\begin{tabular}{ccccgg}
\toprule
\multicolumn{6}{c}{CIFAR-100$-KR$} \\ \cmidrule{1-6}
$D_f(\downarrow)$ & $D_r(\uparrow)$ & $D_{ft}(\downarrow)$ & $D_{rt}(\uparrow)$ & $HM(\uparrow)$ & $HM_{\text{t}}(\uparrow)$ \\
\midrule
0.36 & 97.00 & 0.40 & 92.09 & 98.30 & 95.70 \\
0.98 & 96.73 & 1.80 & 91.87 & 97.86 & 94.93 \\
0.30 & 95.06 & 0.20 & 90.26 & 97.32 & 94.79 \\
1.46 & 96.43 & 1.10 & 91.60 & 97.47 & 95.11 \\
\midrule
1.80 & 96.93 & 1.60 & 92.04 & 97.56 & 95.11 \\
2.10 & 96.71 & 3.50 & 91.93 & 97.30 & 94.16 \\
1.54 & 96.78 & 2.50 & 92.11 & 97.61 & 94.73 \\
1.82 & 96.70 & 2.70 & 92.03 & 97.43 & 94.59 \\
\bottomrule
\end{tabular}
}
\caption{Experiments of incremental unlearning using ViT on CIFAR-100.}
\label{tab:incremental}
\end{table*}
The results of NG and RL indicate significant degradation of the model. 
In the case of NG, as unlearning progresses, the unlearned model tends to converge to a scratch model due to gradient ascent. 
NG requires more unlearning steps in a multi-class unlearning scenario because an increase in the number of classes leads to an increase in the amount of training data. 
Furthermore, when using only forgetting data, NG has no way to preserve the remaining knowledge, causing additional unlearning steps to further degrade the model.
Similarly, RL induces confusion in the model about the forgetting data, potentially leading to unintended outcomes.
This confusion, stemming from the model's attempt to adapt to the randomly assigned new labels, may inadvertently reinforce incorrect associations between features and these labels.  
In multi-class settings, randomly reassigning labels from multiple classes hinders effective unlearning by scattering the model's focus, resulting in the deterioration of the unlearned model.
The impact of BS, is mitigated when the unlearned class is adjacent in feature space, resulting in limited performance in multi-class unlearning.
However, both \methodname{} and \refinemethod{} consistently perform well regardless of the number of forgetting classes.
This is because our methods directly eliminate the knowledge of forgetting data by using modified feature space, thereby minimizing the impact of unlearning steps or the relationships between each forgetting class.

As illustrated in Figure~\ref{fig:multi-class2}, similar to the results presented in Table 1 of our main paper, LAU performs well in multi-class unlearning scenarios.
However, similar to the original model, it suffers from the complete recovery of forgetting knowledge, as shown in KR.
This means that LAU still has enormous forgetting knowledge and it cannot fulfill the \taskname{}.
Conversely, both \methodname{} and \refinemethod{} effectively remove forgetting knowledge, enabling them to perform well in KR.
Although \refinemethod{} slightly improve the forgetting accuracy as growing the number of forgetting classes, they are still reasonable with respect to the forgetting accuracy in KR of retrain model (71.34\% in ViT, 41.28\% in All-CNN).

\subsection{Unlearning with Unseen Data}

To demonstrate the expandability of our methods, we performed unlearning with unseen forgetting data (test data), and the results presented in Table~\ref{tab:unseen}.
Our methods continue to exhibit reasonable performance when utilizing the unlearned model process with forgetting data (seen data) for unlearning.
In terms of accuracy, the gap between using seen data and unseen data is less than 1\%.
Similarly, in KR, this gap remains below 4\%, except for CIFAR-100.
The forget accuracy slightly improved in KR from CIFAR-100, it is also a reasonable result. 
These results indicate that our methods can be applied in more practical scenarios where the original training data is inaccessible.

\begin{figure*}[!t]
    \centering
        \centering
        \includegraphics[width=\textwidth]{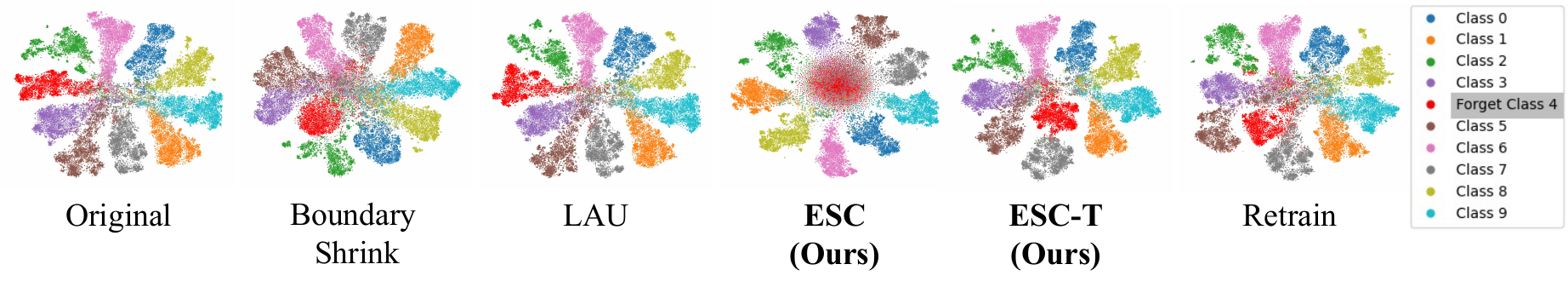}
        \caption{
            t-SNE~\cite{van2008visualizing} visualization of each unlearning method on the CIFAR-10 dataset.
            We removed the knowledge about \textit{'deer'}, which corresponds to class 4, and is represented by a \textcolor{red}{red} dot. 
            Each point represents a sample colored with the ground truth class. 
        }
        \label{fig:tsne}
\end{figure*}
\subsection{Incremental Unlearning with \methodname{} and \refinemethod{}}
In Table~\ref{tab:incremental}, we applied our method repeatedly, i.e. incremental unlearning scenario.
We conducted this using ViT on CIFAR-100 and unlearned 10 classes.
For the sequential setting, we divided the unlearning task as 1, 2, 5, and 10.
For example, in case of 5, we divided 10 classes into 5 subsets and conducted disjoint unlearning for each subset.
Our methods also worked well in incremental setting.
Because each \(U_{P}\) can be merged into \(U_{P_{total}}=\cdots U_{P_2}U_{P_1}\), no additional parameters are needed for incremental unlearning.

\begin{table}[!t]
    \centering
    \resizebox{\columnwidth}{!}{
    \begin{tabular}{lcccccc}
    \toprule
    Method & Forgetting Ratio & 10\% & 20\% & 30\% & 40\% & 50\% \\ \midrule
    \textbf{ESC} & \multirow{2}{*}{\begin{tabular}{c@{}}Avg. Gap \\ with Retrain\end{tabular}($\downarrow$)} & 1.45 & 1.44 & 1.61 & 1.88 & 1.70 \\
    \textbf{ESC-T} & & 2.77 & 2.68 & 3.04 & 2.27 & 3.22 \\
    \bottomrule
    \end{tabular}
}
\caption{Experiments on the impact of various forgetting ratios.}
\label{tab:various_random}
\end{table}
\subsection{Additional Random Data Forgetting}
To demonstrate the robustness of our method to different forgetting ratios, we conducted experiments under various settings.
As shown in Table~\ref{tab:various_random}, our method remains robust across a wide range of forgetting ratios.

\begin{table}[tb]
  \centering
  \resizebox{\columnwidth}{!}{
  \begin{tabular}{@{}lccc@{}}
    \toprule
\multirow{2}{*}{Method}    & All-CNN & \multicolumn{2}{c}{ViT} \\
    \cmidrule(lr){2-2} \cmidrule(lr){3-4}
    & CIFAR-10 & CIFAR-10 & CIFAR-100 \\
    \midrule
    Original & 0.6717 & 0.8265 & 0.9802\\
    Finetune~\cite{warnecke2021machine} & 0.8212 & 0.8939 & 0.9847 \\
    Boundary Shrink~\cite{chen2023boundary} & \underline{0.8719} & 0.9679 & \underline{0.9983} \\
    LAU~\cite{kim2023layer} & 0.7773 & 0.9513 & 0.9869\\
    \methodname (Ours) & \textbf{0.8949} & \textbf{0.9952} & \textbf{0.9989}\\
    \refinemethod (Ours) & 0.8699 & \underline{0.9882} & \textbf{0.9989}\\    
    \midrule
    Retrain & 0.7946 & 0.9378 & 0.9897\\
    \bottomrule
  \end{tabular}
  }
\caption{Zero Retrain Forgetting performance using All-CNN and ViT on CIFAR-10 and CIFAR-100. The larger the value, the closer the unlearned model is to the scratch.}
\label{tab:zrf}
\end{table}
\subsection{Zero Retrain Forgetting}

In addition to the MIA, we further assessed the privacy guarantees of our methods by employing the Zero Retrain Forgetting~(ZRF)~\cite{chundawat2023can}. 
The ZRF calculates the Jensen–Shannon (JS) divergence~\cite{lin1991divergence} between the unlearned model and a random initialized model.
In Table~\ref{tab:zrf}, the ZRF score of our methods is the highest compared to others, reaching almost 0.99 in the ViT experiments.
These results indicate that our methods successfully remove forgetting knowledge in privacy-focused perspective. 
This effectiveness is attributed to our approach's ability to erase knowledge at the feature level.

\begin{figure}[!t]
    \centering
    \includegraphics[width=0.8\columnwidth]{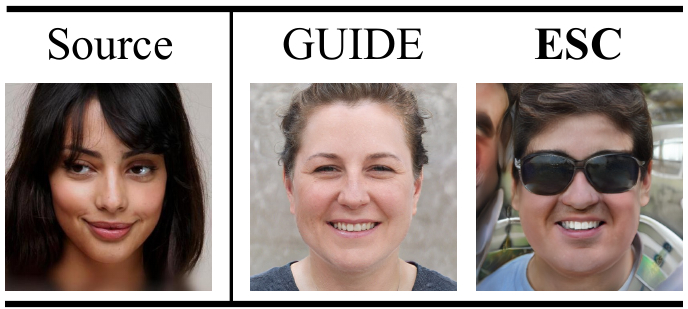} 
    \caption{Qualitative results.}
    \label{fig:gen_qual}
\end{figure}
\begin{table}[!t]
    \centering
    \begin{adjustbox}{width=0.6\columnwidth}
    \begin{tabular}{lcc}
    \toprule
        Method & ID ($\downarrow$) & FID\textsubscript{pre} ($\downarrow$) \\ \hline 
        GUIDE & 0.059 & 7.921 \\
        \textbf{ESC} & \textbf{-0.029} & \textbf{5.690} \\
    \bottomrule
    \end{tabular}
    \end{adjustbox}
    \captionof{table}{Quantitative results.}
    \label{tab:gen_quan}
\end{table}
\subsection{Generative Unlearning}
We also apply ESC to Generative Identity Unlearning (GUIDE)~\cite{seo2024generative}. 
To obtain latent representations, GUIDE uses GOAE\cite{yuan2023make} for inversion and EG3D~\cite{chan2022efficient} as the generative model.
Based on those, GUIDE computes the unlearned latent $w_u$ as below:
\begin{equation}
    w_u=w_0 + w_id,
\end{equation}
where $w_{id}$ is obtained using GOAE.
To remove individual identity while preserving overall generation quality, we apply ESC only to the identity latent feature ($w_{id}$).
As shown in Figure~\ref{fig:gen_qual} and Table~\ref{tab:gen_quan}, the results demonstrate the applicability to generative tasks.

\section{Additional Visualization}
\label{supp:add_vis}

\begin{figure*}[!p]
    \centering
    \includegraphics[width=0.7\textwidth]{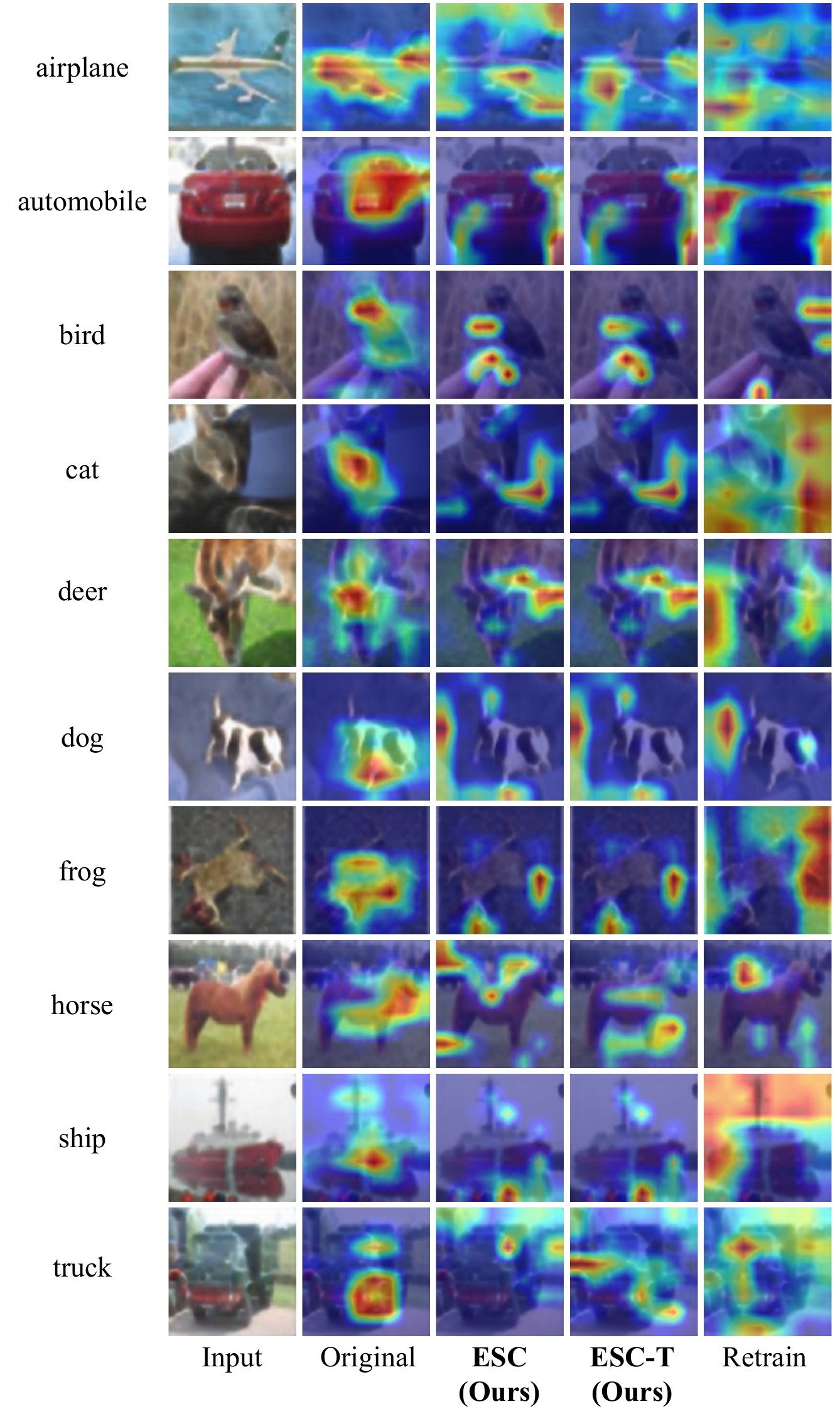}
    \caption{
        The activation map using Grad-CAM on the All-CNN model on the CIFAR-10 dataset.
    }
    \label{fig:gradcam_supp}
\end{figure*}

In this section, we illustrated additional visualization results.
In Figure~\ref{fig:gradcam_supp}, we present the Grad-CAM results for each class in the CIFAR-10 dataset using the All-CNN, complementing the results to Section 4.4. in our main paper.
The findings emphasize once again that both \methodname{} and \refinemethod{} successfully eliminate attention to class object compared with the original model.
Our methods consistently focus strongly on the background regardless of the dataset, indicating that unlearning has been successfully achieved at the feature level.

Furthermore, we utilized t-SNE~\cite{van2008visualizing} for visualizing the manifold space of features, in Figure~\ref{fig:tsne}. 
Upon applying \methodname{}, the clusters for the forgetting class not only centralized among the remaining class clusters but also manifested in a dispersed Gaussian form.
In fact, the activations of the forgetting class are close to zero, leading to these results.
This deactivation also removes the linear separation in the feature space, ensuring the effective deletion of the model's ability to capture distinct features of forgetting knowledge.
When comparing the distributions of Retrain and \refinemethod{}, they appeared quite similar, but \refinemethod{} performed slightly better than Retrain on \metricname{}.
This suggests that the pre-trained information was strongly entangled at the feature level, preventing \taskname{} from being effectively applied through retraining.
This suggests that the pre-trained information is strongly entangled at the feature level, and it means that retrained model is not sufficient for \taskname{}.

\section{Broad Impacts and Limitations}

We reveal a new insight that existing unlearning methods cannot fully fulfill the knowledge removal request by users.
Furthermore, our findings suggest a potential problem: existing methods cannot eliminate almost forgetting knowledge, and we finally suggest a novel perspective for knowledge removal considering user's requests and feature-level, called Knowledge Deletion.
We also propose a novel evaluation setting for this issue.
Our insights provide a benefit to the related research community. 
Furthermore, we introduce simple yet effective methods for \taskname{}.
Our methods demonstrate remarkable performance in various \taskname{} scenarios, including incremental and facial domains, even in the random data forgetting scenario.
In addition to their efficiency, our methods are suitable for real-world AI deployments.

\methodname{} and \refinemethod{} utilize an additional layer after the penultimate layer, i.e., after the feature extractor. 
Our methods are particularly useful when the service provider releases their model as a black-box, such as Chat-GPT~\cite{achiam2023gpt}. 
However, if the model is released as a white-box, we need to integrate this part into the original model architecture.
If we have a simple MLP layer, it can be directly merged with the existing weights, but merging with the entire model remains an open question. 
We expect that this issue could be addressed through methods such as distillation, and we plan to conduct follow-up research on this challenge.

Furthermore, our methods has generalizability and robustness in discriminative tasks, we need to extend this to other domains, such as diffusion model and language model.
Our methods directly edit the feature space, while the latent space of generative models remains highly sensitive.
Consequently, further investigation is needed to determine the applicability of our methods to generative models.
We encourage future research to address these gaps.

\end{document}